# Operations for Learning with Graphical Models

**Wray L. Buntine**                                    WRAY@KRONOS.ARC.NASA.GOV
*RIACS & NASA Ames Research Center, Mail Stop 269–2*
*Moffett Field, CA 94035–1000, USA*

## Abstract

This paper is a multidisciplinary review of empirical, statistical learning from a graphical model perspective. Well-known examples of graphical models include Bayesian networks, directed graphs representing a Markov chain, and undirected networks representing a Markov field. These graphical models are extended to model data analysis and empirical learning using the notation of plates. Graphical operations for simplifying and manipulating a problem are provided including decomposition, differentiation, and the manipulation of probability models from the exponential family. Two standard algorithm schemas for learning are reviewed in a graphical framework: Gibbs sampling and the expectation maximization algorithm. Using these operations and schemas, some popular algorithms can be synthesized from their graphical specification. This includes versions of linear regression, techniques for feed-forward networks, and learning Gaussian and discrete Bayesian networks from data. The paper concludes by sketching some implications for data analysis and summarizing how some popular algorithms fall within the framework presented.

The main original contributions here are the decomposition techniques and the demonstration that graphical models provide a framework for understanding and developing complex learning algorithms.

## 1. Introduction

A probabilistic graphical model is graph where the nodes represent variables and the arcs (directed or undirected) represent dependencies between variables. They are used to define a mathematical form for the joint or conditional probability distribution between variables. Graphical models come in various forms: Bayesian networks used to represent causal and probabilistic processes, data-flow diagrams used to represent deterministic computation, influence diagrams used to represent decision processes, and undirected Markov networks (random fields) used to represent correlation for images and hidden causes.

Graphical models are used in domains such as diagnosis, probabilistic expert systems, and, more recently, in planning and control (Dean & Wellman, 1991; Chan & Shachter, 1992), dynamic systems and time-series (Kjæruff, 1992; Dagum, Galper, Horvitz, & Seiver, 1994), and general data analysis (Gilks et al., 1993a) and statistics (Whittaker, 1990). This paper shows the task of learning can also be modeled with graphical models. This meta-level use of graphical models was first suggested by Spiegelhalter and Lauritzen (1990) in the context of learning probabilities for Bayesian networks.

Graphical models provide a representation for the decomposition of complex problems. They also have an associated set of mathematics and algorithms for their manipulation. When graphical models are discussed, both the graphical formalism and the associated algorithms and mathematics are implicitly included. In fact, the graphical formalism is unnecessary for the technical development of the approach, but its use conveys the important structural information of a problem in a natural visual manner. Graphical operations manipulate the underlying structure of a problem unhindered by the fine detail of the connecting functional and distributional equations. This structuring process is important in the same way that a high-level programming language leads to higher productivity over assembly language.

A graphical model can be developed to represent the basic prediction done by linear regression, a Bayesian network for an expert system, a hidden Markov model, or a connectionist feed-forward network (Buntine, 1994). A graphical model can also be used to represent and reason about the task of *learning* the parameters, weights, and structure of each of these representations. An extension of the standard graphical model that allows this kind of learning to be represented is used here. The extension is the notion of a *plate* introduced by Spiegelhalter[1] (1993). Plates allow samples to be explicitly represented on the graphical model, and thus reasoned about and manipulated. This makes data analysis problems explicit in much the same way that utility and decision nodes are used for decision analysis problems (Shachter, 1986).

This paper develops a framework in which the basic computational techniques for learning can be directly applied to graphical models. This forms the basis of a computational theory of Bayesian learning using the language of graphical models. By a *computational theory* we mean that the approach shows how a wide variety of learning algorithms can be created from graphical specifications and a few simple algorithmic criteria. The basic computational techniques of probabilistic (Bayesian) inference used in this computational theory of learning are widely reviewed (Tanner, 1993; Press, 1989; Kass & Raftery, 1993; Neal, 1993; Bretthorst, 1994). These include various exact methods, Markov chain Monte Carlo methods such as Gibbs sampling, the expectation maximization (EM) algorithm, and the Laplace approximation. More specialized computational techniques also exist for handling missing values (Little & Rubin, 1987), making a batch algorithm incremental, and adapting an algorithm to handle large samples. With creative combination, these techniques are able to address a wide range of data analysis problems.

The paper provides the blueprint for a software toolkit that can be used to construct many data analysis and learning algorithms based on a graphical specification. The conceptual architecture for such a toolkit is given in Figure 1. Probability and decision theory are used to decompose a problem into a computational prescription, and then search and optimization techniques are used to fill the prescription. A version of this toolkit already exists using Gibbs sampling as the general computational scheme (Gilks et al., 1993b). The list of algorithms that can be constructed in one form or another by the scheme in Figure 1 is impressive. But the real gain from the scheme does not arise from the potential re-implementation of existing software, but from understanding gained by putting these in a common language, the ability to create novel hybrid algorithms, and the ability to tailor special purpose algorithms for specific problems.

This paper is tutorial in the sense that it collects material from different communities and presents it in the language of graphical models. This paper introduces graphical models, to represent first-order inference and learning. Second, this paper develops and reviews a number of operations on graphical models. Finally, this paper gives some examples of developing learning algorithms using combinations of the basic operations. The main original contribution is the demonstration that graphical models provide a framework for understanding and developing complex learning algorithms.

---

1. The notion of a "replicated node" was my version of this developed independently. I have adopted the notation of Spiegelhalter and colleagues for uniformity.

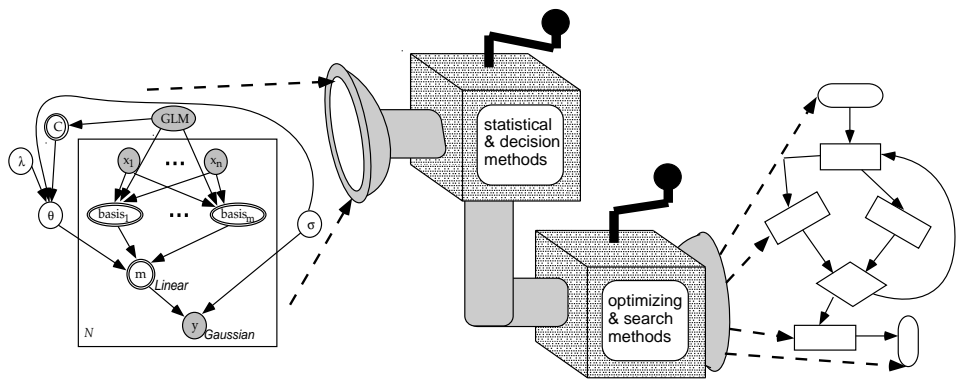

Figure 1: A software generator

In more detail, the paper covers the following topics.

**Introduction to graphical models:**    Graphical models are used in two ways.

> **Graphical models for representing inference tasks:**    Section 2 reviews some basics of graphical models. Graphical models provide a means of representing patterns of inference.

> **Adapting graphical representations to represent learning:**    Section 3 discusses the representation of the problem of learning within graphical models using the notion of a plate.

**Operations on graphical models:**    Operations take a graphical representation of a learning problem and simplify it or perform an important calculation required to solve the problem.

> **Operations using closed-form solutions:**    Section 4 covers those classes of learning problems where closed-form solutions to learning are known. This section adapts standard statistical methods to graphical models.

> **Other basic operations:**    Other operations on graphs are required to be able to handle more complex problems. These are covered in Section 6, including:

> > **Decomposition:** Breaking a learning problem into independent components and evaluating each component.

> > **Differentiation:** Computing the derivative of a probability or log probability with respect to variables on the graph. Differentiation can be decomposed into operations local to groups of nodes in the graph as is popular in neural networks.

> **Some approximate operations:**    Some approximate algorithms follow naturally from the above methods. Section 7 reviews Gibbs sampling and its deterministic cousin, the EM algorithm.

**Some example algorithms:** The closed-form solutions to learning can sometimes be used to form a fast inner loop of more complex algorithms. Section 8 illustrates how graphical models help here.

The conclusion lists some common algorithms and their derivation within the above framework. Proofs of lemmas and theorems are collected in Appendix A.

## 2. Introduction to graphical models

This section introduces graphical models. The brief tour is necessary before introducing the operations for learning.

Graphical models offer a unified qualitative and quantitative framework for representing and reasoning with probabilities and independencies. They combine a representation for uncertain problems with techniques for performing inference. Flexible toolkits and systems exist for applying these techniques (Srinivas & Breese, 1990; Andersen, Olesen, Jensen, & Jensen, 1989; Cowell, 1992). Graphical models are based on the notion of independence, which is worth repeating here.

**Definition 2.1** *A is independent of B given C if $p(A, B|C) = p(A|C)p(B|C)$ whenever $p(C) \neq 0$, for all $A, B, C$.*

The theory of independence as a basic tool for knowledge structuring is developed by Dawid (1979) and Pearl (1988). A graphical model can be equated with the set of probability distributions that satisfy its implied constraints. Two graphical models are *equivalent probability models* if their corresponding sets of satisfying probability distributions are equivalent.

### 2.1 Directed graphical models

The basic kind of graphical model is the Bayesian network, also called belief net, which is most popular in artificial intelligence. See Charniak (1991), Shachter and Heckerman (1987), and Pearl (1988) for an introduction. This is also a graphical representation for a Markov chain. A Bayesian network is a graphical model that uses directed arcs exclusively to form a directed acyclic graph (DAG), (i.e., a directed graph without directed cycles). Figure 2, adapted from (Shachter & Heckerman, 1987) shows a simple Bayesian network for a simplified medical problem. The graphical model represents a conditional decomposition of the joint probability (see Lauritzen, Dawid, Larsen, & Leimer, 1990) for more details and interpretations). This decomposition works as follows (full variable names have been abbreviated).

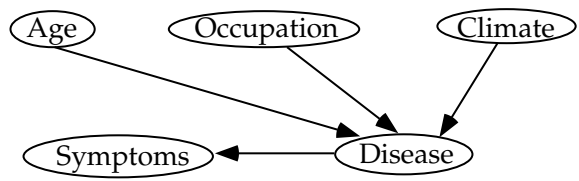

Figure 2: A simplified medical problem

$$p(Age, Occ, Clim, Dis, Symp | M) = \qquad (1)$$
$$p(Age | M) \, p(Occ | M) \, p(Clim | M) \, p(Dis | Age, Occ, Clim, M) \, p(Symp | Dis, M) \,,$$

where $M$ is the conditioning context, for instance the expert's prior knowledge and the choice of the graphical model in Figure 2. Each variable is written conditioned on its *parents*, where $parents(x)$ is the set of variables with a directed arc into $x$. The general form for this equation for a set of variables $X$ is:

$$p(X | M) = \prod_{x \in X} p(x | parents(x), M) \,. \qquad (2)$$

This equation is the *interpretation* of a Bayesian network used in this paper.

## 2.2 Undirected graphical models

Another popular form of graphical model is an undirected graph, sometimes called a Markov network (Pearl, 1988). This is a graphical model for a Markov random field. Markov random fields became used in statistics with the advent of the Hammersley-Clifford theorem (Besag, York, & Mollie, 1991). A variant of the theorem is given later in Theorem 2.1. Markov random fields are used in imaging and spatial reasoning (Ripley, 1981; Geman & Geman, 1984; Besag et al., 1991) and various stochastic models in neural networks (Hertz, Krogh, & Palmer, 1991). Undirected graphs are also important because they simplify the theory of Bayesian networks (Lauritzen et al., 1990).

Figure 3 shows a simple $4 \times 4$ image and an undirected model for the image. This model is based on the first degree Markov assumption; that is, the current pixel is only directly influenced by pixels positioned next to it, as indicated by the undirected arcs between variables $p_{i,j}$ and $p_{i,j+1}$, $p_{i,j}$ and $p_{i+1,j}$, etc. Each node $x$ (corresponding to a pixel) has its set of *neighbors*—those nodes it is directly connected to by an undirected arc. For instance, the neighbors of $p_{1,1}$ are $p_{1,2}$, $p_{2,2}$ and $p_{2,1}$. For the variable/node $x$, denote these by $neighbors(x)$.

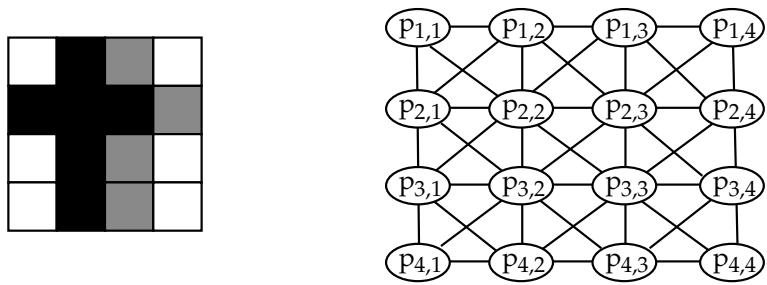

Figure 3: A simple $4 \times 4$ image and its graphical model

In general, there is no formula for undirected graphs in terms of conditional probabilities corresponding to Equation (1) for the Bayesian network of Figure 2. However, a functional decomposition does exist in another form based on the maximal cliques in Figure 3. *Maximal* *cliques* are subgraphs that are fully connected but are not strictly contained in other fully connected subgraphs. These are the 9 sets of $2 \times 2$ cliques such as $\{p_{1,2}, p_{1,3}, p_{2,2}, p_{2,3}\}$. The interpretation of the graph is that the joint probability is a product over functions of the maximal cliques.

$$
\begin{aligned}
p(p_{1,1}, \ldots, p_{4,4}) \ = \ & \qquad\qquad\qquad\qquad\qquad\qquad\qquad\qquad\qquad\qquad (3) \\
& f_1(p_{1,1}, p_{1,2}, p_{2,1}, p_{2,2})\, f_2(p_{1,2}, p_{1,3}, p_{2,2}, p_{2,3})\, f_3(p_{1,3}, p_{1,4}, p_{2,3}, p_{2,4}) \\
& f_4(p_{2,1}, p_{2,2}, p_{3,1}, p_{3,2})\, f_5(p_{2,2}, p_{2,3}, p_{3,2}, p_{3,3})\, f_6(p_{2,3}, p_{2,4}, p_{3,3}, p_{3,4}) \\
& f_7(p_{3,1}, p_{3,2}, p_{4,1}, p_{4,2})\, f_8(p_{3,2}, p_{3,3}, p_{4,2}, p_{4,3})\, f_9(p_{3,3}, p_{3,4}, p_{4,3}, p_{4,4})\ ,
\end{aligned}
$$

for some functions $f_1, \ldots, f_9$ defined up to a constant. From this formula it follows that $p_{1,3}$ is conditionally independent of its non-neighbors given its neighbors $p_{1,2}, p_{2,2}, p_{2,3}, p_{1,4}, p_{2,4}$.

The general form for Equation (3) for a set of variables $X$ is given in the next theorem. Compare this with Equation 2.

**Theorem 2.1** *An undirected graph $G$ is on variables in the set $X$. The set of maximal cliques on $G$ is $Cliques(G) \subset 2^X$. The distribution $p(X)$ (probability or probability density) is strictly positive in the domain $\times_{x \in X} domain(x)$. Then under the distribution $p(X)$, $x$ is independent of $X - \{x\} - neighbors(x)$ given $neighbors(x)$ for all $x \in X$ (Frydenberg (1990) refers to this condition as local G-Markovian) if, and only if, $p(X)$ has the functional representation*

$$
p(X) \ = \ \prod_{C \in Cliques(G)} f_C(C)\ , \qquad\qquad\qquad (4)
$$

*for some functions $f_C > 0$.*

The general form of this theorem for finite discrete domains is called the Hammersley-Clifford Theorem (Geman, 1990; Besag et al., 1991). Again, this equation is used as the *interpretation* of a Markov network.

## 2.3 Conditional probability models

Consider the conditional probability $p(Dis|Age, Occ, Clim)$ found in the simple medical problem from Figure 2 and Equation (1). This conditional probability models how the disease should vary for given values of age, occupation, and climate. Class probability trees (Breiman, Friedman, Olshen, & Stone, 1984; Quinlan, 1992), graphs and rules (Rivest, 1987; Oliver, 1993; Kohavi, 1994), and feed-forward networks are representations devised to express conditional models in different ways. In statistics, the conditional distributions are also represented as regression models and generalized linear models (McCullagh & Nelder, 1989).

The models of Figure 2 and Equation (1) and Figure 3 and Equation (3) show how the joint distribution is composed from simpler components. That is, they give a global model of variables in a problem. The conditional probability models, in contrast, give a model for a subset of variables conditioned on knowing the values of another subset. In diagnosis the concern may be a particular direction for reasoning, such as predicting the disease given patient details and symptoms, so the full joint model provides unnecessary detail. The full joint model may require extra parameters and thus more data to learn. In supervised learning applications, the general view is that conditional models are superior unless prior knowledge dictates a full joint model is more appropriate. This distinction is sometimes referred to as the diagnostic versus the discriminant approach to classification (Dawid, 1976).

There are a number of ways for explicitly representing conditional probability models. Any joint distribution implicitly gives the conditional distribution for any subset of variables, by definition of conditional probability. For instance, if there is a model for $p(Age, Occ, Clim, Dis, Symp)$, then by the definition of conditional probability a conditional model follows:

$$\begin{aligned} p(Dis|Age, Occ, Clim, Symp) &= \frac{p(Age, Occ, Clim, Dis, Symp)}{\sum_{Dis} p(Age, Occ, Clim, Dis, Symp)} \\ &\propto p(Age, Occ, Clim, Dis, Symp) \,. \end{aligned}$$

Conditional distributions can also be represented by a single node that is labeled to identify which functional form the node takes. For instance, in the graphs to follow, labeled Gaussian nodes, linear nodes, and other standard forms are all used. Conditional models such as rule sets and feed-forward networks can be constructed by the use of special deterministic nodes. For instance, Figure 4 shows four model constructs. In each case, the input variables to the conditional models represented have been shaded. This shading means the values of these variables are *known* or given. In Figure 4(b), the shading indicates that the value for $x$ is known, but the value for $c$ is unknown. Presumably $c$ will be predicted using $x$. Figure 4(a) represents a rule set. The nodes with double ovals are *deterministic functions* of their inputs in contrast to the usual nodes, which are probabilistic functions. This means that the value for $rule_1$ is a deterministic function of $x_1, \ldots, x_n$. Notice that this implies that the values for $rule_1$ and the others are known as well. The conditional probability for a deterministic node, as required for Equation (2), is treated as a delta function. Figure 4(c) corresponds to the statement:

$$p(unit|x_1, \ldots, x_n) = \delta_{unit = f(x_1, \ldots, x_n)} = \begin{cases} 1 & \text{if } unit = f(x_1, \ldots, x_n) \,, \\ 0 & \text{otherwise} \end{cases}$$

for some function $f$ not specified. A Bayesian network constructed entirely of double ovals is equivalent to a data flow graph where the inputs are shaded. The analysis of deterministic nodes in Bayesian networks and, more generally, in influence diagrams is considered by Shachter (1990). For some purposes, deterministic nodes are best treated as intermediate variables and removed from the problem. The method for doing this, variable elimination, is given later in Lemma 6.1.

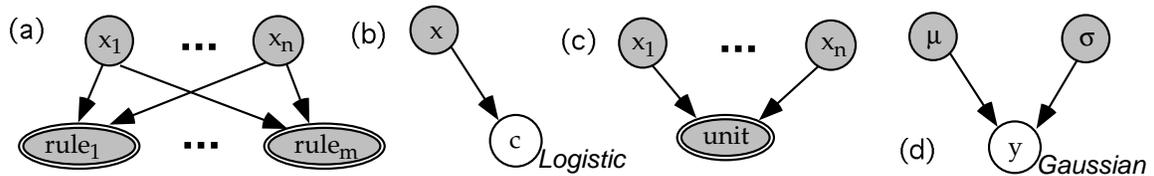

Figure 4: Graphical conditional models

The logical or conjunctive form of each rule in Figure 4(a) is not expressed in the graph, and presumably would be given in the formulas accompanying the graph, however the basic functional structure of the rule set exists. In Figure 4(b), a node has been labeled with its functional type. The functional type for this node with a Boolean variable $c$ is the function,

$$p(c = 1|x) \; = \; \frac{1}{1 + e^x} \; = \; Sigmoid(x) \; = \; Logistic^{-1}(x) \tag{5}$$

which maps a real value $x$ onto a probability in $(0, 1)$ for the binary variable $c$. This function is the inverse of the logistic or logit function used in generalized linear models (McCullagh & Nelder, 1989), and is also the sigmoid function used in feed-forward neural networks. Figure 4(c) uses a deterministic node to reproduce a single unit from a connectionist feed-forward network, where the unit's activation is computed via a sigmoid function. Figure 4(d) is a simple univariate Gaussian, which makes $y$ normally distributed with mean $\mu$ and standard deviation $\sigma$. Here the node is labeled in italics to indicate its conditional type.

At a more general level, networks can be conditional. Figure 5 shows two conditional versions of the simple medical problem. If the shading of nodes is ignored, the joint proba-

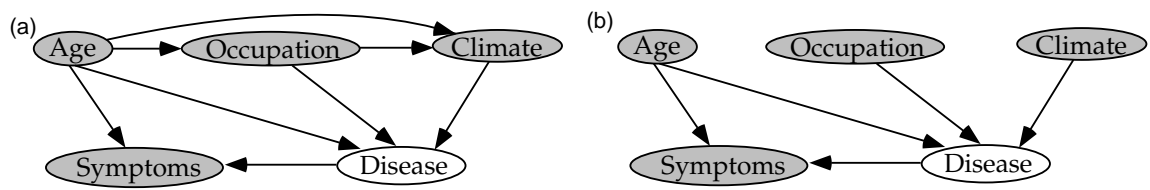

Figure 5: Two equivalent conditional models of the medical problem

bility, $p(Age, Occ, Clim, Dis, Symp)$ for the two graphs (a) and (b) is:

$$p(Age)\, p(Occ|Age)\, p(Clim|Age, Occ)\, p(Dis|Age, Occ, Clim)\, p(Symp|Age, Dis)$$
$$p(Age)\, p(Occ)\, p(Clim)\, p(Dis|Age, Occ, Clim)\, p(Symp|Age, Dis)\, .$$

However, because four of the five nodes are shaded, this means their values are known. The conditional distributions computed from the above are identical:

$$\begin{aligned} &p(Dis|Age, Occ, Clim, Symp) \\ &= \; \frac{p(Dis|Age, Occ, Clim)\, p(Symp|Age, Dis)}{\sum_{Dis} p(Dis|Age, Occ, Clim)\, p(Symp|Age, Dis)}\, . \end{aligned}$$

Why do these distinct graphs become identical when viewed from the conditional perspective? Because conditional components of the model corresponding to age, occupation and climate cancel out when the conditional distribution is formed. However, the symptoms node has the unknown variable disease as a parent, so the arc from age to symptoms is kept.

More generally, the following simple lemma applies and is derived directly from Equation 2.

**Lemma 2.1** *Given a Bayesian network G with some nodes shaded representing a conditional probability distribution, if a node X and all its parents have their values given, then the Bayesian network G′ created by deleting all the arcs into X represents an equivalent probability model to the Bayesian network G.*

This does not mean, for instance in the graphs just discussed, that there is no causal or influential links between the variables age, occupation, and climate, rather that their effects become irrelevant in the conditional model considered because their values are already known. A corresponding result holds for undirected graphs, and follows directly from Theorem 2.1.

**Lemma 2.2** *Given an undirected graph G with some nodes shaded representing a conditional probability distribution, delete an arc between nodes A and B if all their common neighbors are given. The resultant graph G′ represents an equivalent probability model to the graph G.*

## 2.4 Mixed graphical models

Undirected and directed graphs can also be mixed in a sequence. These mixed graphs are called *chain graphs* (Wermuth & Lauritzen, 1989; Frydenberg, 1990). These chain graphs are sometimes used here, However, a precise understanding of them is not required for this paper. A simple chain graph is given in Figure 6. In this case, the single disease node and single symptom node of Figure 2 are expanded to represent the case where there are two possibly co-occurring diseases and three possibly co-occurring symptoms. The medical specialist may have said something like: "Lung disease and heart disease can influence each other, or may have some hidden common cause; however, it is often difficult to tell which is the cause of which, if at all." In the causal model, join the two disease nodes by an undirected arc to represent direct influence. Likewise for the symptom nodes. The resultant joint distribution takes the form:

$$p(Age, Occ, Clim, Heart\text{-}Dis, Lung\text{-}Dis, Symp\text{-}A, Symp\text{-}B, Symp\text{-}C) = \quad (6)$$
$$p(Age)\,p(Occ)\,p(Clim)\,p(Heart\text{-}Dis, Lung\text{-}Dis|Age, Occ, Clim)$$
$$p(Symp\text{-}A, Symp\text{-}B, Symp\text{-}C|Heart\text{-}Dis, Lung\text{-}Dis)$$

where the last two conditional probabilities can take on an arbitrary form. Notice that the probabilities now have more than one variable on the left side.

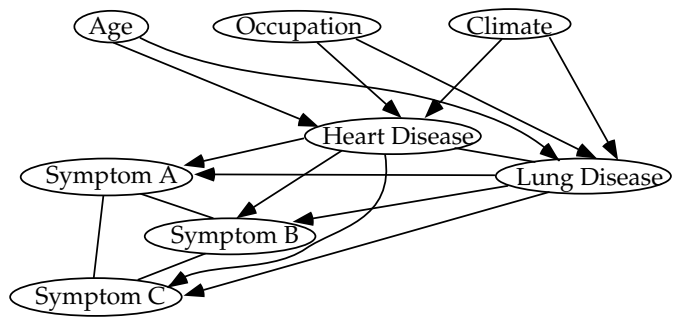

Figure 6: An expanded medical problem

In general, a chain graph consists of a chain of undirected graphs connected by directed arcs. Any cycle through the graph cannot have directed arcs going in opposite directions. Chain graphs can be interpreted as Bayesian networks defined over the components of the chain instead of the original variables. This goes as follows:

**Definition 2.2** *Given a subgraph $G$ over some variables $X$, the chain components are subsets of $X$ that are maximal undirected connected subgraphs in a chain graph $G$ (Frydenberg, 1990). Furthermore, let chain-components$(A)$ denote all nodes in the same chain component as at least one variable in $A$.*

The chain components for the graph above, ordered consistently with the directed arcs, are $\{Age\}$, $\{Occ\}$, $\{Clim\}$, $\{Heart\text{-}Dis, Lung\text{-}Dis\}$, and $\{Symp\text{-}A, Symp\text{-}B, Symp\text{-}C\}$. Informally, a chain graph over variables $X$ with chain components given by the set $T$ is interpreted first as the decomposition corresponding to the decomposition of Bayesian networks in Equation (2):

$$p(X|M) \;=\; \prod_{\tau \in T} p(\tau|parents(\tau), M) \qquad (7)$$

where

$$parents(A) \;=\; \bigcup_{a \in A} parents(a) - A \;.$$

Each component probability $p(\tau|parents(\tau), M)$ has a form similar to Equation (4).

Sometimes, to process graphs of this form without having to consider the mathematics of chain graphs, the following device is used.

**Comment 2.1** *When a set of nodes $U$ in a chain graph form a clique (a fully connected subgraph), and all have identical children and parents otherwise, then the set of nodes can be replaced a single node representing the cross product of the variables.*

This operation for Figure 6 is done to get Figure 7. Furthermore, chain graphs are sometimes

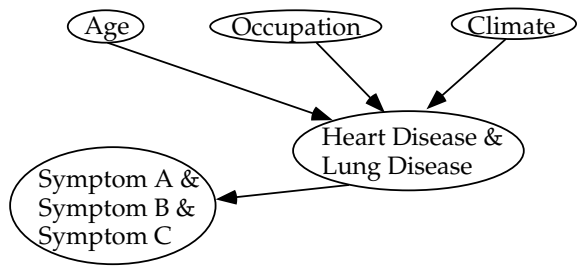

Figure 7: An expanded medical problem

used here where Bayesian networks can also be used. In this case:

**Comment 2.2** *The chain components of a Bayesian network are the singleton sets of individual variables in the graph. Furthermore, chain-components$(A) = A$.*

Chain graphs can be decomposed into a chain of directed and undirected graphs. An example is given in Figure 8. Figure 8(a) shows the original chain graph. Figure 8(b) shows its directed and undirected components together with the Bayesian network on the right showing how they are pieced together. Having done this decomposition, the components are analyzed using all the machinery of directed and undirected graphs. The interpretation of these graphs in terms of independence statements and the implied functional form of the joint probability is a combination of the previous two forms given in Equation (2) and Theorem 2.1, based on (Frydenberg, 1990, Theorem 4.1), and on the interpretation of conditional graphical models in Section 2.3.

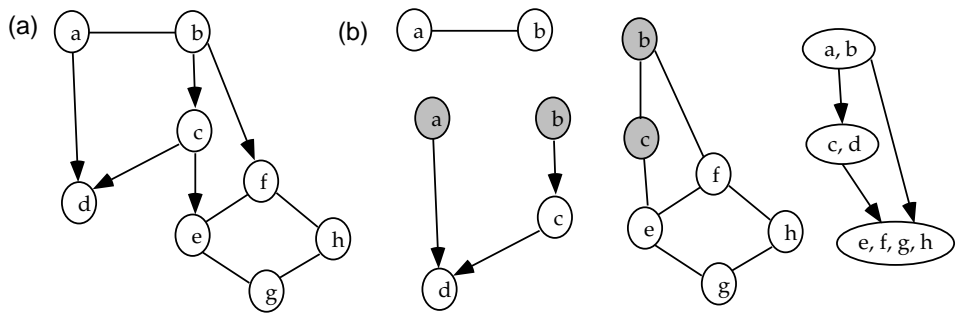

Figure 8: Decomposing a chain graph

## 3. Introduction to learning with graphical models

A simplified inference problem is represented in Figure 9. Here, the nodes $var_1$, $var_2$ and $var_3$ are shaded. This represents that the value of these nodes is given, so the inference task is to predict the value of the remaining variable *class*. This graph matches the so-called "idiot's" Bayes classifier (Duda & Hart, 1973; Langley, Iba, & Thompson, 1992) used in supervised learning for its speed and simplicity. The probabilities on this network are easily learned from data about the three input variables $var_1$, $var_2$ and $var_3$, and *class*. This graph also matches an unsupervised learning problem where the class *class* is not in the data but is hidden. An unsupervised learning algorithm learns hidden classes (Cheeseman, Self, Kelly, Taylor, Freeman, & Stutz, 1988; McLachlan & Basford, 1988).

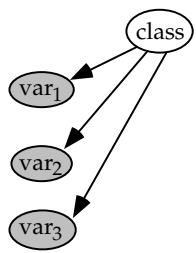

Figure 9: A simple classification problem

The implied joint for these variables read from the graph is:

$$p(class, var_1, var_2, var_3) = p(class) \, p(var_1|class) \, p(var_2|class) \, p(var_3|class) \ . \quad (8)$$

The Bayesian classifier gets its name because it is derived by applying Bayes theorem to this joint to get the conditional formula:

$$p(class|var_1, var_2, var_3) = \frac{p(class)p(var_1|class) \, p(var_2|class) \, p(var_3|class)}{\sum_{class} p(class)p(var_1|class) \, p(var_2|class) \, p(var_3|class)} \ . \quad (9)$$

The same formula is used to predict the hidden class for objects in the simple unsupervised learning framework. Again, this formula, and corresponding formula for more general classifiers, can be found automatically by using exact methods for inference on Bayesian networks.

Consider the simple model given in Figure 9. If the matching unsupervised learning problem for this model was represented, a sample of $N$ cases of the variables would be observed, with the first case being $var_{1,1}$, $var_{2,1}$, $var_{3,1}$, and the $N$-th case being $var_{1,N}$, $var_{2,N}$, $var_{3,N}$. The corresponding hidden classes, $class_1$ to $class_N$, would not be observed, but interest would be in performing inference about the parameters needed to specify the hidden classes. The learning problem is represented in Figure 10. This includes two added features: an explicit representation of the model parameters $\phi$ and $\theta$, and a representation of the sample as $N$ repeated subgraphs. The parameter $\phi$ (a vector of class probabilities) gives the proportions for the hidden classes, and the three parameters $\theta_1$, $\theta_2$ and $\theta_3$ give how the variables are distributed within each hidden class. For instance, if there are 10 classes, then $\phi$ is a vector of 10 class probabilities such that the prior probability of a case being in class $c$ is $\phi_c$. If $var_1$ is a binary variable, then $\theta_1$ would be 10 probabilities, one for each class, such that if the case is known to be in class $c$, then the probability $var_1$ is true is given by $\theta_{1,c}$ and the probability $var_1$ is false is given by $1 - \theta_{1,c}$. This yields the following equations:

$$p(class = c|\phi, M) = \phi_c \ ,$$
$$p(var_j = true|class = c, \theta_j, M) = \theta_{j,c} \ .$$

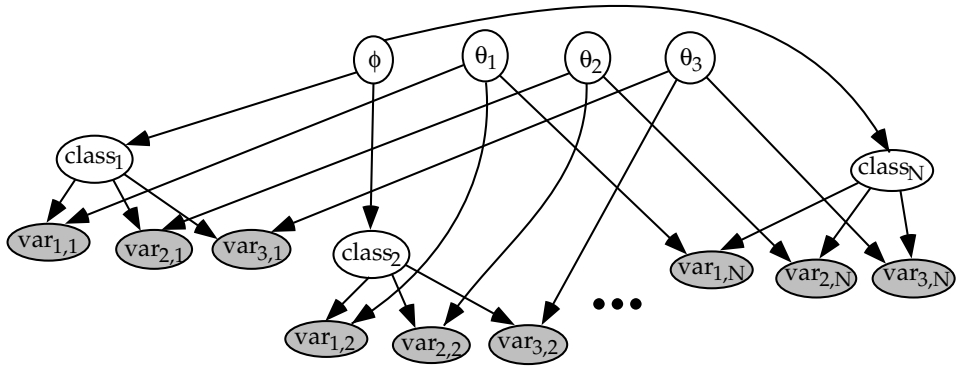

Figure 10: Learning the simple classification

The unknown model parameters $\phi$, $\theta_1$, $\theta_2$ and $\theta_3$ are included in the graphical model to explicitly represent all unknown variables and parameters in the learning problem.

## 3.1 Introduction to Bayesian learning

Now is a useful time to introduce the basic terminology of Bayesian learning theory. This is not an introduction to the field. Introductions are given in (Bretthorst, 1994; Press, 1989; Loredo, 1992; Bernardo & Smith, 1994; Cheeseman, 1990). This section reviews notions such as the sample likelihood and Bayes factor, important for subsequent results.

For the above unsupervised learning problem there is the *model*, $M$, which is the use of the hidden class and the particular graphical structure of Figure 10. There are data assumed to be independently sampled, and there are the *parameters* of the model ($\phi$, $\theta_1$, etc.). In order to use the theory, it must be assumed that the model is correct. That is, the "true" distribution for the data can be assumed to come from this model with some parameters. In practice, hopefully the model assumptions are sufficiently close to the truth. Different sets of model assumptions may be tried. Typically, the "true" model parameters are unknown, although there may be some rough idea about their values. Sometimes, several models are considered (for instance different kinds of Bayesian networks), but it is assumed that just one of them is correct. Model selection or model averaging methods are used to deal with them.

For the Bayesian classifier above, a subjective probability is placed over the model parameters, in the form $p(\phi, \theta_1, \theta_2, \theta_3|M)$. This is called the *prior probability*. Bayesian statistics and decision theory is distinguished from all other statistical approaches in that it places initial probability distributions, the prior probability, over unknown model parameters. If the model is a feed-forward neural network, then a prior probability needs to be placed over the network weights and the standard deviation of the error. If the model is linear regression with Gaussian error, then it is over the linear parameters $\theta$ and the standard deviation of the error. Prior probabilities are an active area of research and are discussed in most introductions to Bayesian methods.

The next important component is the sample *likelihood*, which, on the basis of the model assumptions $M$ and given a set of parameters $\phi$, $\theta_1$, $\theta_2$ and $\theta_3$, says how likely the sample of data was. This is $p(sample|\phi, \theta_1, \theta_2, \theta_3, M)$. The model needs to completely determine the sample likelihood. The sample likelihood is the basis of the maximum likelihood principle and many hypothesis testing methods (Casella & Berger, 1990). This combines with the prior to form the *posterior probability*:

$$p(\phi, \theta_1, \theta_2, \theta_3|sample, M) \;=\; \frac{p(sample|\phi, \theta_1, \theta_2, \theta_3, M)p(\phi, \theta_1, \theta_2, \theta_3|M)}{p(sample|M)} \; .$$

This equation is Bayes theorem and the term $p(sample|M)$ is derived from the prior and sample likelihood using an integration or sum that is often difficult to do:

$$p(sample|M) \;=\; \int_{\phi, \theta_1, \theta_2, \theta_3} p(sample|\phi, \theta_1, \theta_2, \theta_3, M)p(\phi, \theta_1, \theta_2, \theta_3|M) \, \mathrm{d}(\phi, \theta_1, \theta_2, \theta_3) \; . \quad (10)$$

This term is called the *evidence* for model $M$, or *model likelihood*, and is the basis for most Bayesian model selection, model averaging methods, and Bayesian hypothesis testing methods using Bayes factors (Smith & Spiegelhalter, 1980; Kass & Raftery, 1993). The *Bayes factor* is a relative quantity used to compare one model $M_1$ with another $M_2$:

$$Bayes\text{-}factor(M_2, M_1) \; = \; \frac{p(sample|M_2)}{p(sample|M_1)} \; .$$

Kass and Raftery (1993) review the large variety of methods available for computing or estimating the evidence for a model including numerical integration, importance sampling, and the Laplace approximation. In implementation, the log of the Bayes factor is used to keep the arithmetic within reasonable bounds. The log of the evidence can still produce large numbers, and since rounding errors in floating point arithmetic scales with the order of magnitude, the log Bayes factor is the preferred quantity to consider in implementation. The evidence is often simpler in mathematical analysis. The Bayes factor is the Bayesian equivalent to the likelihood ratio test used in orthodox statistics and developed by Wilks. See Casella and Berger (1990) for an introduction and Vuong (1989) for a recent review.

The evidence and Bayes factors are fundamental to Bayesian methods. It is often the case that a complex "non-parametric" model (a statistical term that loosely translates as "many and varied parameter" model) be used for a problem, rather than a simple model with some fixed number of parameters. Examples of such models are decision trees, most neural networks, and Bayesian networks. For instance, suppose two models are proposed with $M_1$ and $M_2$ being two Bayesian networks suggested by the domain expert. These are given in Figure 11. They are over two multinomial variables $var_1$ and $var_2$ and two Gaussian variables $x_1$ and $x_2$. Model $M_2$ has an additional arc going from the discrete variable $var_2$ to the real valued variable $x_1$.

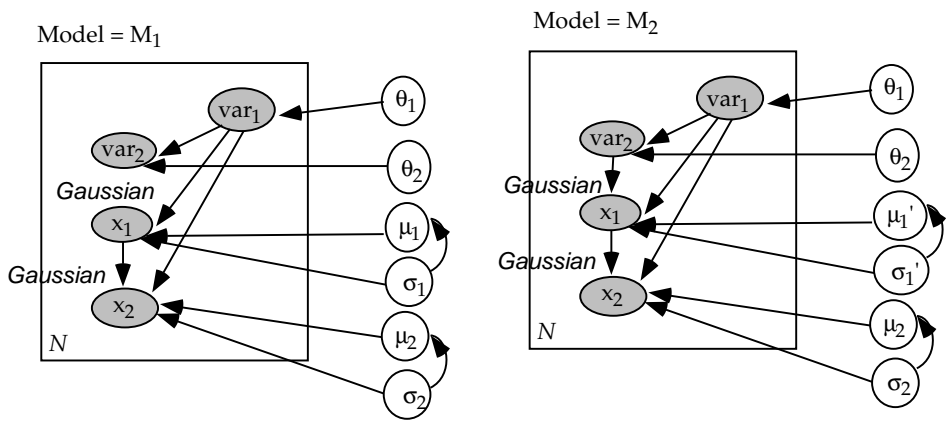

Figure 11: Two graphical models. Which should learning select?

The parameters $\theta_1, \theta_2, \mu_1, \sigma_1, \mu_2, \sigma_2$ for model $M_1$ parameterize probability distributions for the first Bayesian network, and the parameters $\theta_1, \theta_2, \mu'_1, \sigma'_1, \mu_2, \sigma_2$ for the second. The task is to learn not only a set of parameters, but also to select a Bayesian network from the two. The Bayes factor gives the comparative worth of the two models. This simple example extends in principle to selecting a single decision tree, rule set, or Bayesian network from the huge number available from attributes in the domain. In this case compare the posterior probabilities of the two models $p(M_1|sample)$ and $p(M_2|sample)$. Assuming the truth falls in one or other model, the first is computed using Bayes theorem as:

$$
\begin{aligned}
p(M_1|sample) &= \frac{p(sample|M_1)p(M_1)}{p(sample|M_1)p(M_1) + p(sample|M_2)p(M_2)} \;, \\
&= \frac{1}{1 + Bayes\text{-}factor(M_2, M_1)\frac{p(M_2)}{p(M_1)}} \;.
\end{aligned}
$$

More generally, when multiple models exist, it still holds that:

$$
\frac{p(M_2|sample)}{p(M_1|sample)} \;=\; Bayes\text{-}factor(M_2, M_1)\frac{p(M_2)}{p(M_1)}.
$$

Notice that the computation requires of each model its prior and its evidence. The second form reduces the computation to the relative quantity being the Bayes factor, and a ratio of the priors. Bayesian hypothesis testing corresponds to checking if the Bayes factor of the null hypothesis compared to the alternative hypothesis is very small or very large. Bayesian model building corresponds to searching for a model with a high value of $p(M|sample) \propto p(sample|M)p(M)$, which usually involves comparing Bayes factors of this model with alternative models during the search.

When making an estimate about a new case $x$, the estimate becomes:

$$
\begin{aligned}
&p(x|sample, \{M_1, M_2\}) \\
&= p(M_1|sample)\,p(x|sample, M_1) + p(M_2|sample)\,p(x|sample, M_2) \\
&= \frac{p(sample|M_1)\,p(M_1)\,p(x|sample, M_1) + p(sample|M_2)\,p(M_2)\,p(x|sample, M_2)}{p(sample|M_1)\,p(M_1) + p(sample|M_2)\,p(M_2)} \;.
\end{aligned}
$$

The predictions of the individual models is averaged according to the model posteriors $p(M_1|sample)$ and $p(M_2|sample) = 1 - p(M_1|sample)$. The general components used in this calculation are the model priors, the evidence for each model or the Bayes factors, and the prediction for the new case made for each model.

This process of *model averaging* happens in general. A typical non-parametric problem would be to learn class probability trees from data. The number of class probability tree models is super-exponential in the number of features. Even when learning Bayesian networks from data the number of Bayesian networks is at best quadratic in the number of features. Doing an exhaustive search of these spaces and doing the full averaging implied by the equation above is computationally infeasible in general. It may be the case that 15 models have posterior probabilities $p(M|sample)$ between 0.1 and 0.01, and several thousand more models have posteriors from 0.001 to 0.0000001. Rather than select a single model, a representative set of several models might be chosen and averaged using the identity:

$$
p(x|sample) \;=\; \sum_i p(M_i|sample)\,p(x|sample, M_i) \;.
$$

The general averaging process is depicted in Figure 12 where a Gibbs sampler is used to generate a representative subset of models with high posterior. This kind of computation is done for class probability trees where representative sets of trees are found using a heuristic branch and bound algorithm (Buntine, 1991b), and for learning Bayesian networks (Madigan & Raftery, 1994). A sampling scheme for Bayesian networks is presented in Section 8.3.

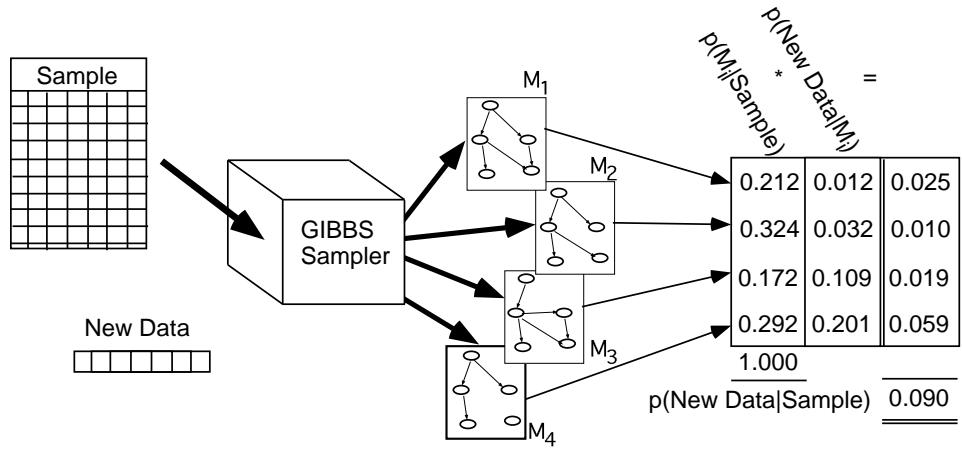

Figure 12: Averaging over multiple Bayesian networks

## 3.2 Plates: representing learning with graphical models

In their current form, graphical models do not allow the convenient representation of a learning problem. There are four important points to be observed regarding the use of graphical models to improve their suitability for learning. Consider again the unsupervised learning system described in the introduction of Section 3.

- The unknown model parameters $\phi$, $\theta_1$, $\theta_2$, and $\theta_3$ are included in the graphical model to explicitly represent all variables in the problem, even model parameters. By including these in the probabilistic model, an explicitly Bayesian model is constructed. Every variable in a graphical model, even unknown model parameters, has a defined prior probability.

- The learning sample is a repeated set of measured variables so the basic model of Figure 9 appears duplicated as many times as there are cases in the sample, as shown in Figure 10. Clearly, this awkward repetition will occur whenever homogeneous data is being modeled (typical in learning). Techniques for handling this repetition form a major part of this paper.

- Neither graph in Figures 9 and 10 represents the goal of learning. For learning to be goal directed, additional information needs to be included in the graph: how is learned knowledge evaluated or how can subsequent performance be measured? This is the role of decision theory and it is modeled in graphical form using influence diagrams (Shachter, 1986). This is not discussed here, but is covered in (Buntine, 1994).

- Finally, it must be possible to take a graphical representation of a learning problem and the goal of learning and construct an algorithm to solve the problem. Subsequent sections discuss techniques for this.

Consider a simplified version of the same unsupervised problem. In fact, the simplest possible learning problem containing uncertainty goes as follows: there is a biased coin with

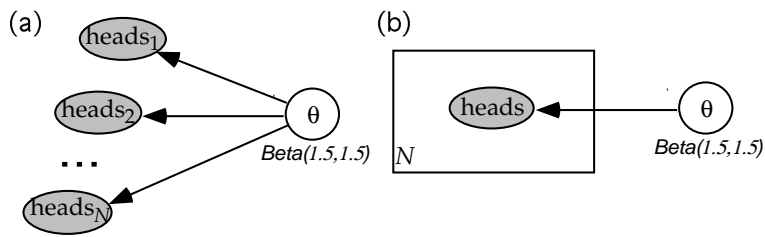

Figure 13: Tossing a coin: model without and with a plate

an unknown bias for heads $\theta$. That is, the long-run frequency of getting heads for this coin on a fair toss is $\theta$. The coin is tossed $N$ times and each time the binary variable $heads_i$ is recorded. The graphical model for this is in Figure 13(a). The $heads_i$ nodes are shaded because their values are given, but the $\theta$ node is not. The $\theta$ node has a $Beta(1.5, 1.5)$ prior. This assumes $\theta$ is distributed according to the Beta distribution with parameters $\alpha_1 = 1.5$ and $\alpha_2 = 1.5$,

$$p(\theta|\alpha_1, \alpha_2) = \frac{\theta^{\alpha_1 - 1}(1 - \theta)^{\alpha_2 - 1}}{Beta(\alpha_1, \alpha_2)} \qquad (11)$$

where $Beta(,)$ is the standard beta function given in many mathematical tables. This prior is plotted in Figure 14. A $Beta(1.0, 1.0)$ prior, for instance, is uniform in $\theta$, whereas $Beta(1.5, 1.5)$ slightly favors values closer to 0.5—a fairer coin. Figure 13(b) is an equivalent

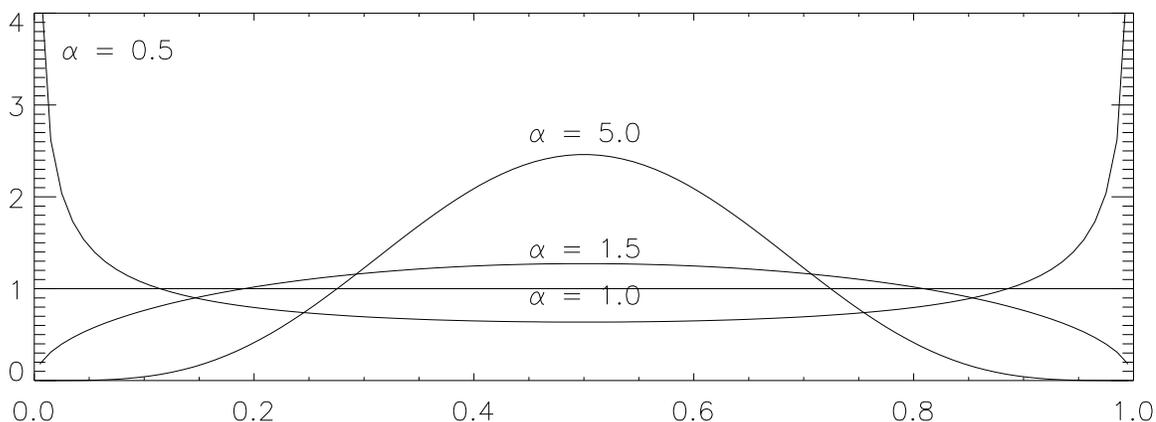

Figure 14: The $Beta(1.5, 1.5)$ prior ($\alpha = 1.5$) and other priors on $\theta$

graphical model using the notation of plates. The repeated group, in this case the $heads_i$ nodes, is replaced by a single node with a box around it. The box is referred to as a *plate*. and implies that

- the enclosed subgraph is duplicated $N$ times (into a "stack" of plates),

- the enclosed variables are indexed, and

- any exterior-interior links are duplicated.

In Section 2.1 it was shown that any Bayesian network has a corresponding form for the joint probability of variables in the Bayesian network. The same applies to plates. The plate indicates that a product ($\prod$) will appear in the corresponding form. The probability equation for Figure 10, read directly from the graph, is:

$$p(\phi, \theta_1, \theta_2, \theta_3, class_1, var_{1,1}, var_{2,1}, var_{3,1}, \ldots, class_N, var_{1,N}, var_{2,N}, var_{3,N}) =$$
$$p(\phi)\,p(\theta_1)\,p(\theta_2)\,p(\theta_3)\,p(class_1|\phi)\,p(var_{1,1}|class_1, \theta_1)\,p(var_{2,1}|class_1, \theta_2)\,p(var_{3,1}|class_1, \theta_3)$$
$$\ldots\, p(class_N|\phi)\,p(var_{1,N}|class_N, \theta_1)\,p(var_{2,N}|class_N, \theta_2)\,p(var_{3,N}|class_N, \theta_3)\,.$$

The corresponding equation using product notation is:

$$p(\phi, \theta_1, \theta_2, \theta_3, class_i, var_{1,i}, var_{2,i}, var_{3,i} : i = 1, \ldots, N) = p(\phi)\,p(\theta_1)\,p(\theta_2)\,p(\theta_3)$$
$$\prod_{i=1}^{N} p(class_i|\phi)\,p(var_{1,i}|class_i, \theta_1)\,p(var_{2,i}|class_i, \theta_2)\,p(var_{3,i}|class_i, \theta_3)\,.$$

These two equations are equivalent. However, the differences in their written form corresponds to the differences in their graphical form. Each plate is converted into a product: the joint probability ignoring the plates is written, a product ($\prod$) is added for each plate to index the variables inside it. If there are two disjoint plates then there are two disjoint products. Overlapping plates yield overlapping products. The corresponding transformation for the unsupervised learning problem of Figure 10 is given in Figure 15. Notice, the hidden *class* variable is not shaded so it is not given. The corresponding transformation for the supervised learning problem of Figure 10, where the classes are given, and thus corresponds to the idiot's Bayes classifier, is identical to Figure 15 except that the *class* variable is shaded because the classes are now part of the training sample.

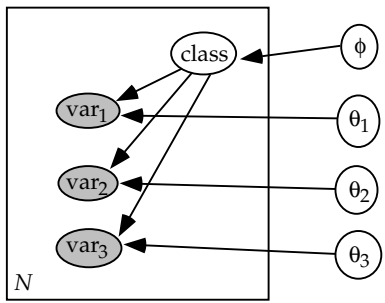

Figure 15: Simple unsupervised learning, with a plate

Many learning problems can be similarly modeled with plates. Write down the graphical model for the full learning problem with only a single case provided. Put a box around the data part of the model, pull out the model parameters (for instance, the weights of the network or the classification parameters), and ensure they are unshaded because they are unknown. Now add the data set size ($N$) to the bottom left corner.

The notion of a plate is formalized below. This formalization is included for use in subsequent proofs.

**Definition 3.1** *A chain graph $G$ with plates on variable set $X$ consists of a chain graph $G'$ on variables $X$ with additional boxes called plates placed around groups of variables. Only directed arcs can cross plate boundaries, and plates can be overlapping. Each plate $P$ has an integer $N_P$ in the bottom left corner indicating its cardinality. Each plate indexes the variables inside it with values $i = 1, \ldots, N_P$. Each variable $V \in X$ occurs in some subset of the plates. Let $indval(V)$ denote the set of values for indices corresponding to these plates. That is, $indval(V)$ is the cross product of index sets $\{1, \ldots, N_P\}$ for plates $P$ containing $V$.*

A graph with plates can be expanded to remove the plates. Figure 10 is the expanded form of Figure 15. Given a chain graph with plates $G$ on variables $X$, construct the *expanded graph* as follows:

- For each variable $V \in X$, add a node for $V_i$ for each $i \in indvar(V)$.

- For each undirected arc between variables $U$ and $V$, add an undirected arc between $U_i$ and $V_i$ for $i \in indvar(V) = indvar(U)$.

- For each directed arc between variables $U$ and $V$, add a directed arc between $U_i$ and $V_j$ for $i \in indvar(V)$ and $j \in indvar(V)$ where $i$ and $j$ have identical values for index components from the same plate.

The parents for indexed variables in a graph with plates are the parents in the expanded graph.

$$parents(U) = \bigcup_{i \in indval(U)} parents(U_i) \,.$$

A graph with plates is interpreted using the following product form. If the product form for the chain graph $G'$ without plates with chain components $T$ is

$$p(X|M(G')) = \prod_{\tau \in T} p(\tau | parents(\tau), M) \,,$$

then the product form for the chain graph $G$ with plates has a product for each plate:

$$p(X|M(G)) = \prod_{\tau \in T} \prod_{i \in indval(\tau)} p(\tau_i | parents(\tau_i), M) \,. \tag{12}$$

This is given by the expanded version of the graph. Testing for independence on chain graphs with plates involves expanding the plates. In some cases, this can be simplified.

## 4. Exact operations on graphical models

This section introduces basic inference methods on graphs without plates and exact inference methods on graphs with plates. While there are no common machine learning algorithms explained in this section, the operations explained are the mathematical basis of most fast learning algorithms. Therefore, the importance of these basic operations should not be underestimated. Their use within more well-known learning algorithms is explained in later sections.

Once a graphical model is developed to represent a problem, the graph can be manipulated using various exact or approximate transformations to simplify the problem. This section reviews basic exact transformations available: arc reversal, node removal, and exact removal of plates by recursive arc reversal. The summary of operations emphasizes the computational aspects. A graphical model has an associated set of definitions or tables for the basic functions and conditional probabilities implied by the graph, the operations given below effect both the graphical structure and these underlying mathematical specifications. In both cases, the process of making these transformations should be constructive so that a graphical specification for a learning problem can be converted into an algorithm.

There are several generic approaches for performing inference on directed and undirected networks without plates. These approaches are mentioned, but will not be covered in detail. The first approach is exact and corresponds to removing independent or irrelevant information from the graph, then attempting to optimize an exact probabilistic computation by finding a reordering of the variables. The second approach to performing inference is approximate and corresponds to approximate algorithms such as Gibbs sampling, and other Markov chain Monte Carlo methods (Hrycej, 1990; Hertz et al., 1991; Neal, 1993). In some cases, the complexity of the first approach is inherently exponential in the number of variables, so the second can be more efficient. The two approaches can be combined in some cases after appropriate reformulation of the problem (Dagum & Horvitz, 1992).

## 4.1 Exact inference without plates

The exact inference approach has been highly refined for the case where all variables are discrete. It is not surprising that available algorithms have strong similarities (Shachter, Andersen, & Szolovits, 1994) since the major choice points involve the ordering of the summation and whether this ordering is selected dynamically or statically. Other special classes of inference algorithms include the cases where the model is a multivariate Gaussian (Shachter & Kenley, 1989; Whittaker, 1990), or corresponds to some specific diagnostic structure, such as two-level believe networks with a level of symptoms connected to a level of diseases (Henrion, 1990). This subsection reviews some simple, exact transformations on graphical models without plates. Two representative methods are covered but are by no means optimal: arc reversal and arc removal. They are important, however, because they are the building blocks on which methods for graphs with plates are based. Many more sophisticated variations and combinations of these algorithms exist in the literature, including the handling of deterministic nodes (Shachter, 1990) and chain graphs and undirected graphs (Frydenberg, 1990).

### 4.1.1 ARC REVERSAL

Two basic steps for inference are to marginalize nuisance parameters or to condition on new evidence. This may require evaluating probability variables in a different order. The *arc reversal* operator interchanges the order of two nodes connected by a directed arc (Shachter, 1986). This operator corresponds to Bayes theorem and is used, for instance, to automate the derivation of Equation (9) from Equation (8). The operator applies to directed acyclic graphs and to chain graphs where $a$ and $b$ are adjacent chain components.

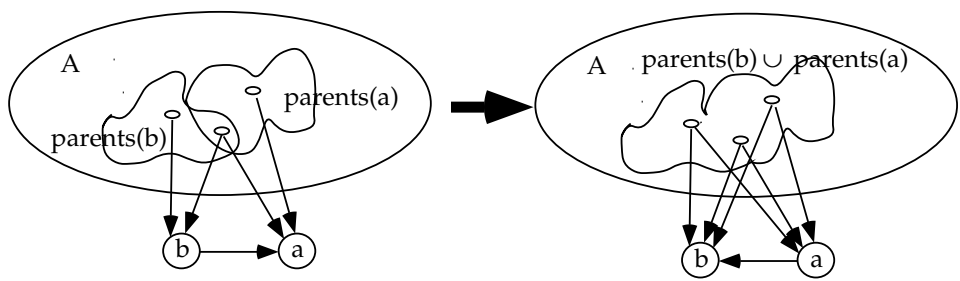

Figure 16: Arc reversal: reversing nodes $a$ and $b$

Consider a fragment of a graphical model as given in the left of Figure 16. The equation for this fragment is:

$$p(a, b | A) = p(b | parents(b)) \, p(a | b, parents(a)) \, .$$

Suppose nodes $a$ and $b$ need to be reordered. Assume that between $a$ and $b$ there is no directed path of length greater than one. If there was, then reversing the arc between $a$ and $b$ would create a cycle, which is forbidden in a Bayesian network and chain graph. The formula for the variable reordering can be found by applying Bayes theorem to the above equation.

$$p(a | A) = \sum_b p(b | parents(b)) \, p(a | parents(a)) \, ,$$

$$p(b | a, A) = \frac{p(b | parents(b)) \, p(a | parents(a))}{\sum_b p(b | parents(b)) \, p(a | parents(a))} \, .$$

The corresponding graph is given in the right of Figure 16. Notice that the effect on the graph is that nodes for $a$ and $b$ now share their parents. This is an important point. If all of $a$'s parents were also $b$'s, and vice versa, excepting $b$ ($parents(a) = parents(b) \cup \{b\}$), then the graph would be unchanged except for the direction of the arc between $a$ and $b$. Regardless, the probability tables or formula associated with the graph also need to be updated. If the variables are discrete and full conditional probability tables are maintained, then this operation requires instantiating the set $\{a, b\} \cup parents(a) \cup parents(b)$ in all ways, which is exponential in the number of variables.

### 4.1.2 Arc and node removal

Some variables in a graph are part of the model, but are not important for the goal of the data analysis. These are called *nuisance parameters*. An unshaded node $y$ without children (no outward going arcs) that is neither an action node nor a utility node can always be removed from a Bayesian network. This corresponds to leaving out the term $p(y | parents(y))$ in the product of Equation 2. Given

$$p(a, b, y) = p(a) \, p(b | a) \, p(y | a, b)$$

then $y$ can be *marginalized* out trivially to yield:

$$p(a, b) = p(a) \, p(b | a) \, .$$

More generally this applies to chain graphs—a chain component whose nodes are all unshaded and have no children can be removed. If $y$ is a node without children, then remove the node with $y$ from the graph and the arcs to it; ignore the factor $p(y|parents(y))$ in the full joint form. Consider that the $i$-th case in Figure 10 (nodes $class_i, var_{1,i}, var_{2,i}, var_{3,i}$) can be removed from the model without affecting the rest of the graph.

## 4.2 Removal of plates by exact methods

Consider the simple coins problem of Figure 13 again. The graph represents the joint probability for $p(\theta, heads_1, \ldots, heads_N)$. The main question of interest here is the conditional probability of $\theta$ given the data $heads_1, \ldots, heads_N$. This could be obtained through repeated arc reversals between $\theta$ and $heads_1$, then between $\theta$ and $heads_2$, and so on, until all the data appears before $\theta$ in the directed graph. Doing this repeated series of applications of Bayes theorem yields a fully connected graph with $(N+1)N/2$ arcs. The corresponding formula for the posterior simplified with Lemma 2.1 is also simple:

$$p(\theta|heads_1, \ldots, heads_N, \alpha_1 = 1.5, \alpha_2 = 1.5) = \frac{\theta^{\alpha_1 - 1 + p}(1-\theta)^{\alpha_2 - 1 + n}}{Beta(\alpha_1 + p, \alpha_2 + n)} \tag{13}$$

where $p$ is the number of heads in the sequence and $n = N - p$ is the number of tails. This is a worthwhile introductory exercise in Bayesian decision theory (Howard, 1970) that should be familiar to most students of statistics. Compare this with Equation (11). There are several important points to notice about this result:

- Effectively, this does a parameter update, $\alpha_1' = \alpha_1 + p$ and $\alpha_2' = \alpha_2 + n$, requiring no search or numerical optimization. The whole sequence of tosses, irrespective of its length and the ordering of the heads and tails, can be summed up with two numbers. These summary statistics are called *sufficient statistics* because, assuming the model used is correct, they are sufficient to explain all that is important about $\theta$ in the data,.

- The corresponding graph can be simplified as shown in Figure 17. The plate is efficiently removed and replaced by the sufficient statistics (two numbers) irrespective of the size of the sample.

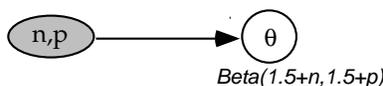

Figure 17: Removing the plate in the coin problem

- The posterior distribution has a simple form. Furthermore, all the moments of $\theta$, $\log \theta$, and $\log(1-\theta)$ for the distribution can be computed as simple functions of the normalizing constant, $Beta(\alpha_1', \alpha_2')$. For instance,

$$\mathcal{E}_{\theta|heads_1, \ldots, heads_N, \alpha_1, \alpha_2}(\log \theta) = \frac{\partial \log Beta(\alpha_1', \alpha_2')}{\partial \alpha_1},$$

$$\mathcal{E}_{\theta|heads_1, \ldots, heads_N, \alpha_1, \alpha_2}(\theta) = \frac{Beta(\alpha_1' + 1, \alpha_2')}{Beta(\alpha_1', \alpha_2')},$$

$$\mathcal{E}_{\theta|heads_1,\dots,heads_N,\alpha_1,\alpha_2}\left(\left(\theta-\overline{\theta}\right)^2\right) \;=\; \frac{Beta(\alpha_1'+2,\alpha_2')}{Beta(\alpha_1',\alpha_2')} - \frac{Beta^2(\alpha_1'+1,\alpha_2')}{Beta^2(\alpha_1',\alpha_2')} \;.$$

This result might seem somewhat obscure, but it is a general property holding for a large class of distributions that allows some averages to be calculated by symbolic manipulation of the normalizing constant.

### 4.3 The exponential family

This result generalizes to a much larger class of distributions referred to as the exponential family (Casella & Berger, 1990; DeGroot, 1970). This includes standard undergraduate distributions such as Gaussians, Chi squared, and Gamma, and many more complex distributions constructed from simple components including class probability trees over discrete input domains (Buntine, 1991b), simple discrete and Gaussian versions of a Bayesian network (Whittaker, 1990), and linear regression with a Gaussian error. Thus, these are a broad and not insignificant class of distributions that are given in the definition below. Their general form has a linear combination of parameters and data in the exponential.

**Definition 4.1** *A space $X$ is* independent *of the parameter $\theta$ if the space remains the same when just $\theta$ is changed. If the domains of $x, y$ are independent of $\theta$, then the conditional distribution for $x$ given $y$, $p(x|y,\theta,M)$, is in the* exponential family *when*

$$p(x|y,\theta,M) \;=\; \frac{h(x,y)}{Z(\theta)}\exp\left(\sum_{i=1}^{k} w_i(\theta)t_i(x,y)\right) \tag{14}$$

*for some functions $w_i$, $t_i$, $h$ and $Z$ and some integer $k$, for $h(x,y) > 0$. The normalization constant $Z(\theta)$ is known as the* partition function.

Notice the functional form of Equation (14) is similar to the functional form for an undirected graph of Equation (4), as holds in many cases for a Markov random field. For the previous coin tossing example, both the coin tossing distribution (a binomial on $heads_i$) and the posterior distribution on the model parameters ($\theta$) are in the exponential family. To see this, notice the following rewrites of the original probabilities. These make the components $w_i$, $t_i$ and $Z$ explicit.

$$p(heads|\theta) \;=\; \exp\left(1_{heads=true}\log\theta + 1_{heads=false}\log(1-\theta)\right)\;,$$
$$p(\theta|heads_1,\dots,heads_N,\alpha_1,\alpha_2) \;=\;$$
$$\frac{1}{Beta(\alpha_1+p,\alpha_2+n)}\exp\left((\alpha_1+p-1)\log\theta + (\alpha_2+n-1)\log(1-\theta)\right)\;.$$

Table 2 in Appendix B gives a selection of distributions, and their functional form. Further details can be found in most textbooks on probability distributions (DeGroot, 1970; Bernardo & Smith, 1994).

The following is a simple graphical reinterpretation of the Pitman-Koopman Theorem from statistics (Jeffreys, 1961; DeGroot, 1970). In Figure 18(a), $T(x_*,y_*)$ is a statistic of fixed dimension independent of the sample size $N$ (corresponding to $n, p$ in the coin tossing example). The theorem says that the sample in Figure 18(a) can be summarized in statistics, as shown in Figure 18(b), if and only if the probability distribution for $x|y,\theta$ is in the exponential family. In this case, $T(x_*,y_*)$ is a sufficient statistic.

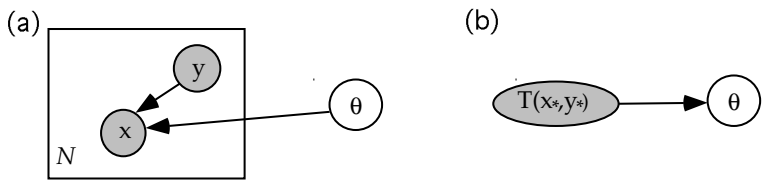

Figure 18: The generalized graph for plate removal

**Theorem 4.1** (Recursive arc-reversal). *Consider the model $M$ represented by the graphical model for a sample of size $N$ given in Figure 18(a). Have $x$ in the domain $X$ and $y$ in the domain $Y$, both domains are independent of $\theta$, and both domains have components that are real valued or finite discrete. Let the conditional distribution for $x$ given $y, \theta$ be $f(x|y, \theta)$, which is positive for all $x \in X$. If first derivatives exist with respect to all real valued components of $x$ and $y$, the plate removal operation applies for all samples $x_* = x_1, \ldots, x_N$, $y_* = y_1, \ldots, y_N$, and $\theta$, as given in Figure 18(b), for some sufficient statistics $T(x_*, y_*)$ of dimension independent of $N$ if and only if the conditional distribution for $x$ given $y, \theta$ is in the exponential family, given by Equation (14). In this case, $T(x_*, y_*)$ is an invertible function of the $k$ averages:*

$$\frac{1}{N} \sum_{j=1}^{N} t_i(x_j) \quad : \quad i = 1, \ldots, k \ .$$

In some cases, this extends to domains $X$ and $Y$ dependent on $\theta$ (Jeffreys, 1961).

Sufficient statistics for a distribution from the exponential family are easily read from the functional form by taking a logarithm. For instance, for the multivariate Gaussian, the sufficient statistics are $x_i$ for $i = 1, \ldots, d$ and $x_i x_j$ for $0 \le i \le j \le d$, and the normalizing constant $Z(\mu, \Sigma)$ is given by:

$$Z(\mu, \Sigma) \ = \ \frac{(2\pi)^{d/2}}{\det^{1/2} \Sigma} \exp\left(\frac{1}{2} \mu^\dagger \Sigma \mu\right) \ .$$

As for coin tossing, it generally holds that if a sampling distribution (a binomial on $heads_i$) is in the exponential family, then the posterior distribution for the model parameters ($\theta$) can also be cast as exponential family. This is only useful when the normalizing constant (this is $Beta(\alpha_1 + p, \alpha_2 + n)$ in the coin tossing example) and its derivatives are readily computed.

**Lemma 4.1** (The conjugacy property). *In the context in Theorem 4.1, assume the distribution for $x$ given $y, \theta$ can be represented by the exponential family. Factor the normalizing constant $Z(\theta)$ into two components, $Z(\theta) = Z_1(\theta) Z_2$, where the second is the constant part independent of $\theta$. Assume the prior on $\theta$ takes the form:*

$$p(\theta|\tau, M) \ = \ \frac{f(\theta)}{Z_\theta(\tau)} \exp\left(\tau_{k+1}(\log 1/Z_1(\theta)) + \sum_{i=1}^{k} \tau_i w_i(\theta)\right) \tag{15}$$

for some $k+1$ dimensional parameter $\tau$, where $Z_\theta(\tau)$ is the appropriate normalizing constant and $f(\theta)$ is any function. Then the posterior distribution for $\theta$, $p(\theta|\tau, x_1, \ldots, x_N, M)$, is also represented by Equation (15) with the parameters

$$
\begin{aligned}
\tau'_{k+1} &= \tau_{k+1} + N \\
\tau'_i &= \tau_i + \sum_{j=1}^{N} t_i(x_j, y_j) \qquad i = 1, \ldots, k .
\end{aligned}
$$

When the function $f(\theta)$ is trivial, for instance uniformly equal to 1, then the distribution in Equation (15) is referred to as the *conjugate* distribution, which means it has a mathematical form mirroring that of the sample likelihood. The prior parameters $\tau$, by looking at the update equations in the lemma, can be thought of as corresponding to the sufficient statistics from some "prior sample" and given by $\tau_{k+1}$.

This property is useful for analytic and computational purposes. Once the posterior distribution is found, and assuming it is one of the standard distributions, the property can easily be established. Table 3 in Appendix B gives some standard conjugate prior distributions for those in Table 2, and Table 4 gives their matching posteriors. More extensive summaries of this are given by DeGroot (1970) and Bernardo and Smith (1994). The parameters for these priors can be set using standard reference priors (Box & Tiao, 1973; Bernardo & Smith, 1994) or elicited from a domain expert.

There are several other important consequences of the Pitman-Koopman Theorem or recursive arc reversal that should not go unnoticed.

**Comment 4.1** *If $x, y$ are discrete and finite valued, then the distribution $p(x|y, \theta)$ can be represented as a member of the exponential family. This holds because a positive finite discrete distribution can always be represented as an extended case statement in the form*

$$
p(x|y, \theta) = \exp\left( \sum_{i=1,k} 1_{t_i(x,y)} f_i(\theta) \right)
$$

*where the boolean functions $t_i(x, y)$ are a set of mutually exclusive and exhaustive conditions. The indicator function $1_A$ has the value 1 if the boolean $A$ is true and 0 otherwise. The main importance of the exponential family is in continuous or integer domains. Of course, since a large class of functions $\log p(x|y, \theta)$ can always be approximated arbitrarily well by a polynomial in $x, y$ and $\theta$ with sufficiently many terms, the exponential family covers a broad class of distributions.*

The application of the exponential family to learning is perhaps the earliest published result on computational learning theory. The following two interpretations of the recursive arc reversal theorem are relevant for distributions involving continuous variables.

**Comment 4.2** *An incremental learning algorithm with finite memory must compress the information it has seen so far in the training sample into a smaller set of statistics. This can only be done without sacrificing information in the sample, in a context where all probabilities are positive, if the hypothesis or search space of learning is a distribution from the exponential family.*

**Comment 4.3** *The computational requirements for learning an exponential family distribution are guaranteed to be linear in the sample size: first compute the sufficient statistics and then learning proceeds independently of the sample size. This could be exponential in the dimension of the feature space, however.*

Furthermore, in the case where the functions $w_i$ are *full rank* in $\theta$ (dimension of $\theta$ is $k$, same as $w$, and the Jacobian of $w$ with respect to $\theta$ is invertible, $\det\left(\frac{\mathrm{d}w(\theta)}{\mathrm{d}\theta}\right) \neq 0$), various moments of the distribution can easily be found. For this situation, the function $w^{-1}$, when it exists, is called the *link function* (McCullagh & Nelder, 1989).

**Lemma 4.2** *Consider the notation of Definition 4.1. If the link function $w^{-1}$ for an exponential family distribution exists, then moments of functions of $t_i(x, y)$ and $\exp(t_i(x, y))$ can be expressed in terms of derivatives and direct applications of the functions $Z$, $t_i$, $w_i$, and $w^{-1}$. If the normalizing constant $Z$ and the link function are in closed form, then so will the moments.*

Techniques for doing these symbolic calculations are given in Appendix B. Exponential family distributions then fall into two groups. There are those where the normalizing constant and link function are known, such as the Gaussian. One can efficiently compute their moments and determine the functional form of their conjugate distributions up to the normalizing constant. For others this is not the case. For others, such as a Markov random field used in image processing, moments can generally only be computed by an approximation process like Gibbs sampling given in Section 7.1.

### 4.4 Linear regression: an example

As an example, consider the problem of linear regression with Gaussian error described in Figure 19. This is an instance of a generalized linear model and has a linear construction

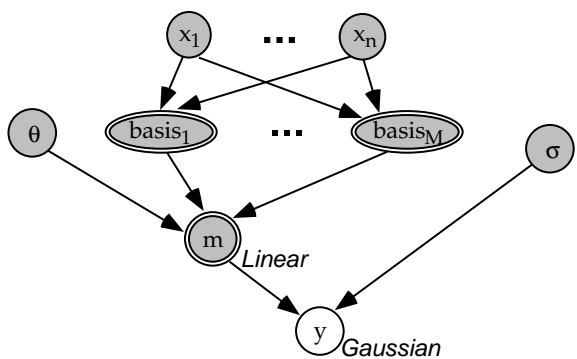

Figure 19: Linear regression with Gaussian error

$$m = \sum_{j=1}^{M} \theta_j basis_j(x_1, \ldots, x_n)$$

at its core. The $M$ basis functions are known deterministic functions of the input variables $x_1, \ldots, x_n$. These would typically be nonlinear orthogonal functions such as Legendre polynomials. These combine linearly with the parameters $\theta$ to produce the mean $m$ for the Gaussian.

The corresponding learning problem represented as plates is expressed in Figure 20. The

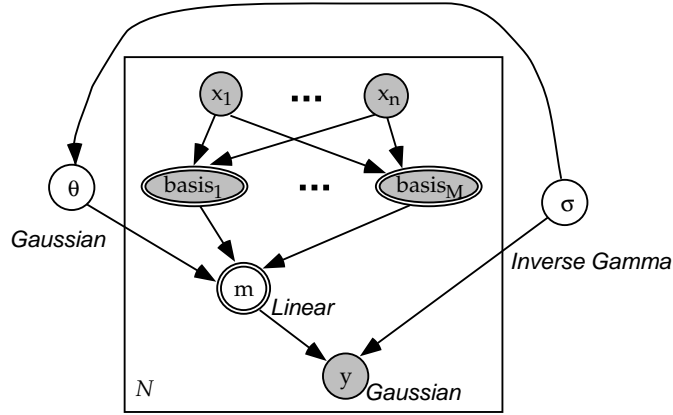

Figure 20: The linear regression problem

joint probability for this model is as follows:

$$p(\sigma)\, p(\theta|\sigma) \frac{1}{\left(\sqrt{2\pi}\sigma\right)^N} e^{-\frac{1}{2\sigma^2} \sum_{i=1}^N \left(y_i - \sum_{j=1}^M \theta_j\, basis_j(x_{\cdot,i})\right)^2}$$

where $x_{\cdot,i}$ denotes the vector of values for the $i$-th datum, $x_{1,i}, \ldots, x_{n,i}$. Linear regression with Gaussian error falls in the exponential family because a Gaussian is in the exponential family and the mean of the simple Gaussian is a linear function of the regression parameters (see Lemma 5.1).

For this case, the correspondence to the exponential family is drawn as follows. The individual data likelihoods, $p(y|x_1, \ldots, x_n, \theta, \sigma)$, need only be considered. Expand the probability to show it is a linear sum of data terms and parameter terms.

$$
\begin{aligned}
& p(y|x_1, \ldots, x_n, \theta, \sigma) \\
={} & \frac{1}{\sqrt{2\pi}\sigma} \exp\left(-\frac{1}{2\sigma^2}\left(y - \sum_{j=1}^M basis_j(x_{\cdot})\, \theta_j\right)^2\right), \\
={} & \frac{1}{\sqrt{2\pi}\sigma} \exp\left(-\frac{1}{2\sigma^2} y^2 - \sum_{j,k=1}^M basis_j(x_{\cdot})\, basis_k(x_{\cdot}) \frac{\theta_j \theta_k}{2\sigma^2} + \sum_{j=1}^M basis_j(x_{\cdot})\, y\, \frac{\theta_j}{2\sigma^2}\right).
\end{aligned}
$$

The data likelihood in this last line can be seen to be in the same form as the general exponential family where the sufficient statistics are the various data terms in the exponential. Also, the link function does not exist because there are $M$ parameters and $M(M+1)/2$ sufficient statistics.

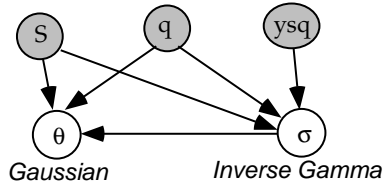

Figure 21: The linear regression problem with the plate removed

The model of Figure 20 can therefore be simplified to the graph in Figure 21, where $q$ and $S$ are the usual sample means and covariances obtained from the so-called normal equations of linear regression. $S$ is a matrix of dimension $M$ (the number of basis functions) and $q$ is a vector of dimension $M$.

$$S_{j,k} = \frac{1}{N} \sum_{i=1}^{N} basis_j(x_{.,i}) \, basis_k(x_{.,i})$$

$$q_j = \frac{1}{N} \sum_{i=1}^{N} basis_j(x_{.,i}) \, y_i$$

$$ysq = \frac{1}{N} \sum_{i=1}^{N} y_i^2 \, .$$

These three sufficient statistics can be read directly from the data likelihood above.

Consider the formula:

$$\int_y p(y|x_1, \ldots, x_n, \theta, \sigma) \, \mathrm{d}y \, .$$

Differentiating with respect to $\theta_i$ shows that the expected value of $y$ given $x_1, \ldots, x_n$ and $\theta, \sigma$ is the mean (as expected):

$$\mathcal{E}_{y|x_1, \ldots, x_n, \theta, \sigma}(y) = m = \sum_{j=1}^{M} basis_j(x_.) \, \theta_j \, .$$

Differentiating with respect to $\sigma$ shows that the expected error from the mean is $\sigma^2$ (again expected):

$$\mathcal{E}_{y|x_1, \ldots, x_n, \theta, \sigma}\left((y - m)^2\right) = \sigma^2 \, .$$

Higher-order derivatives give formula for higher-order moments such as skewness and kurtosis (Casella & Berger, 1990), which are functions of the second, third and fourth central moments. While these are well known for the Gaussian, the interesting point is that these formula are constructed by differentiating the component functions in Equation (14) without recourse to integration. Finally, the conjugate distribution for the parameters $\theta$ in this linear regression problem is the multivariate Gaussian distribution when $\sigma$ is known, and for $\sigma^2$ is the inverted Gamma.

## 5. Recognizing and using the exponential family

How can recursive arc reversal be applied automatically to a graphical model? First, when a graphical model or some subset of a graphical model falls in the exponential family needs to be identified. If each conditional distribution in a Bayesian network or chain component in a chain graph is exponential family, then the full joint is exponential family. The following lemma gives this with some additional conditions for deterministic nodes. This applies to Bayesian networks using Comment 2.2.

**Lemma 5.1** *A chain graph has a single plate. Let the non-deterministic variables inside the plate be $X$, and the deterministic variables be $Y$. Let the variables outside the plate be $\theta$. If:*

1. *All arcs crossing a plate boundary are directed into the plate.*

2. *For all chain components $\tau$, the conditional distribution $p(\tau | parents(\tau))$ is from the exponential family with data variables from $(X, Y)$ and model parameters from $\theta$; furthermore $\log p(\tau | parents(\tau), \theta)$ is a polynomial function of variables in $Y$.*

3. *Each variable $y \in Y$ can be expressed as a deterministic function of the form*

$$y = \sum_{i=1}^{l} u_i(X) \, v_i(\theta)$$

*for some functions $u_i, v_i$.*

*Then the conditional distribution $p(X, Y | \theta)$ is from the exponential family.*

Second, how can these results be used when the model does not fall in the exponential family? There are two categories of techniques available in this context. In both cases, the algorithms concerned can be constructed from the graphical specifications. The two new classes together with the recursive arc reversal case are given in Figure 22. In each case, (I) denotes the graphical configuration and (II) denotes the operations and simplifications performed by the algorithm. When the various normalization constants are known in closed form and appropriate moments and Bayes factors can be computed quickly, all three algorithm schemas have reasonable computational properties.

The first category is where a useful subset of the model does fall into the exponential family. This is represented by the *partial exponential family* in Figure 22. The part of the problem that is exponential family is simplified using the recursive arc reversal of Theorem 4.1, and the remaining part of the problem is typically handled approximately. Decision trees and Bayesian networks over multinomial or Gaussian variables also fall into this category. This happens because when the structure of the tree or Bayesian network is given the remaining problem is composed of a product of multinomials or Gaussians. This is the basis of various Bayesian algorithms developed for these problems (Buntine, 1991b; Madigan & Raftery, 1994; Buntine, 1991c; Spiegelhalter, Dawid, Lauritzen, & Cowell, 1993; Heckerman, Geiger, & Chickering, 1994). Strictly speaking, decision trees and Bayesian networks over multinomial or Gaussian variables are in the exponential family (see Comment 4.1). However, it is more computationally convenient to treat them this way. This category is discussed more in Section 8.

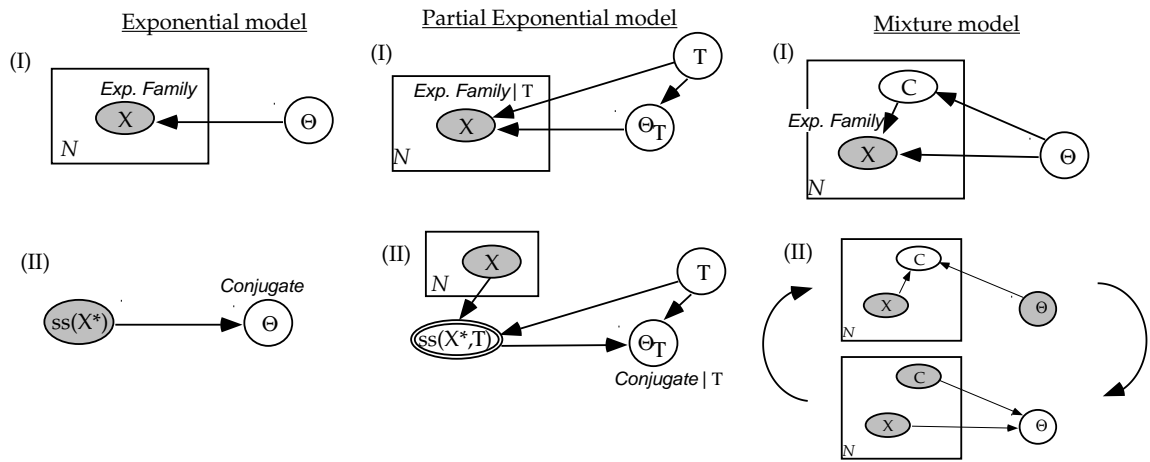

Figure 22: Three categories of algorithms using the exponential family

The second category is where, if some hidden variables are introduced into the data, the problem becomes exponential family if the hidden values are known. This is represented by the *mixture model* in Figure 22. Mixture models (Titterington, Smith, & Makov, 1985; Poland, 1994) are used to model unsupervised learning, incomplete data in the classification problems, robust regression, and general density estimation. Mixture models extend the exponential family to a rich class of distributions, so this second category is an important one in practice. General methods for handling these problems correspond to Gibbs sampling (and other Markov chain Monte Carlo methods) discussed in Section 7.2 and its deterministic counterpart the expectation maximization algorithm, discussed in Section 7.4. As shown in Figure 22, these algorithms cycle back and forth between a process that re-estimates $c$ given $\theta$ using first-order inference and a process that uses the fast exponential family algorithms to re-estimate $\theta$ given $c$.

## 6. Other operations on graphical models

The recursive arc reversal theorem of Section 4.2 characterizes when plates can be readily removed and the sample summarized in some statistics. Outside of these cases, more general classes of approximate algorithms exist. Several of these are introduced in Section 7 and more detail is given, for instance, by Tanner (1993). These more general algorithms require a number of basic operations be performed on graphs:

**Decomposition:** A learning problem can sometimes be decomposed into simpler subproblems with each yielding to separate analysis. One form of decomposition of learning problems is considered in Section 6.2. Another related form that applies to undirected graphs is developed by Dawid and Lauritzen (1993). Other forms of decomposition can be done at the modeling level, where the initial model is constructed in a manner requiring fewer parameters, as is Heckerman's similarity networks (1991).

**Exact Bayes factors:** Model selection and averaging methods are used to deal with multiple models (Kass & Raftery, 1993; Buntine, 1991b; Stewart, 1987; Madigan & Raftery, 1994). These require the computation of Bayes factors for models constructed during search. Exact methods for computing Bayes factors are considered in Section 6.3.

**Derivatives:** Various approximation and search algorithms require derivatives be calculated, as discussed next.

## 6.1 Derivatives

An important operation on graphs is the calculation of derivatives of parameters. This is useful after conditioning on the known data to do approximate inference. Numerical optimization using derivatives can be done to search for MAP values of parameters, or to apply the Laplace approximation to estimate moments. This section shows how to compute derivatives using operations local to each node. The computation is therefore easily parallelized, as is popular, for instance, in neural networks.

Suppose a graph is used to compile a function that searches for the MAP values of parameters in the graph conditioned on the known data. In general, this requires use of numerical optimization methods (Gill, Murray, & Wright, 1981). To use a gradient descent, conjugate gradient or Levenberg-Marquardt approach requires calculation of first derivatives. To use a Newton-Raphson approach requires calculation of second derivatives, as well. While this could be done numerically by difference approximations, more accurate calculations exist. Methods for symbolically differentiating networks of functions, and piecing together the results to produce global derivatives are well understood (Griewank & Corliss, 1991). For instance, software is available for taking a function defined in Fortran, C++ code, or some other language, to produce a second function that computes the exact derivative. These problems are also well understood for feed-forward networks (Werbos, McAvoy, & Su, 1992; Buntine & Weigend, 1994), and graphical models with plates only add some additional complexity. The basic results are reproduced in this section and some simple examples given to highlight special characteristics arising from their use with chain graphs.

Consider the problem of learning a feed-forward network. A simple feed-forward network is given in Figure 23(a). The corresponding learning problem is given in Figure 23(b), representing the feed-forward network as a Bayesian network. Here the sigmoid units of the network are modeled with deterministic nodes, and the network output represents the mean of a bivariate Gaussian with inverse variance matrix $\Sigma$. Because of the nonlinear sigmoid function making the deterministic mapping from inputs $x_1, x_2, x_3$ to the means $m_1, m_2$, this learning problem has no reasonable component falling in the exponential family. A rough fallback method is to calculate a MAP value for the weight parameters. This would be the method used for the Laplace approximation (Buntine & Weigend, 1991; MacKay, 1992) covered in (Tanner, 1993; Tierney & Kadane, 1986). The setting of priors for feed-forward networks is difficult (MacKay, 1993; Nowlan & Hinton, 1992; Wolpert, 1994), and it will not be considered here other than assuming a prior is used, $p(w)$. The graph implies the

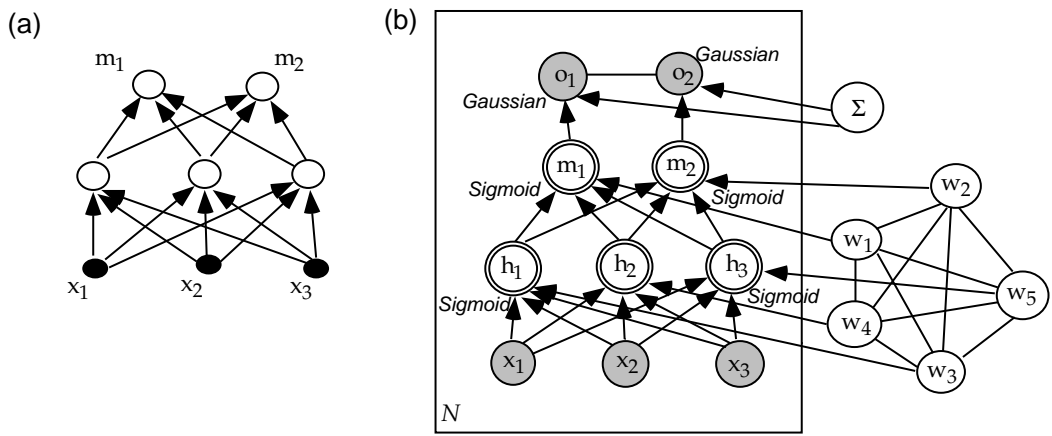

Figure 23: Learning a feed-forward network

following posterior probability:

$$p(\Sigma, w_1, \ldots, w_5 \mid o_{1,i}, o_{2,i}, x_{1,i}, x_{2,i}, x_{3,i} \; : \; i = 1, \ldots, N) \qquad (16)$$

$$\propto \quad p(\Sigma) \, p(w_1, \ldots, w_5) \prod_{i=1}^{N} \frac{\det^{1/2} \Sigma}{2\pi} \exp\left(\frac{1}{2}(o_i - m_i)^\dagger \Sigma (o_i - m_i)\right)$$

$$m_i = Sigmoid(w_i^\dagger h)$$

$$h_i = Sigmoid(w_{i+2}^\dagger x) \; .$$

The undirected clique on the parameters $w$ indicates the prior has a term $p(w)$. Suppose the posterior is differentiated with respect to the parameters $w_4$. The result is well known to the neural network community since this kind of calculation yields the standard back-propagation equations.

Rather than work through this calculation, instead look at the general case. To develop the general formula for differentiating a graphical model, a few more concepts are needed. Deterministic nodes form islands of determinism within the uncertainty represented by the graph. Partial derivatives within each island can be calculated via recursive use of the chain rule, for instance, by forward or backward propagation of derivatives through the equations. For instance, *forward propagation* for the above network gives:

$$\frac{\partial m_1}{\partial w_4} = \sum_{i=1}^{3} \frac{\partial m_1}{\partial h_i} \frac{\partial h_i}{\partial w_4} \; .$$

This is called forward propagation because the derivatives with respect to $w_4$ are propagated forward in the network. In contrast, *backward propagation* would propagate derivatives of $m_1$ with respect to different variables backwards. For each island of determinism, the important variables are the output variables, and their derivatives are required. So for the feed-forward network above, partial derivatives of $m_1$, $m_2$, and $m_3$ with respect to $w_4$ are required.

**Definition 6.1** *The* non-deterministic children *of a node $x$, denoted $ndchildren(x)$, are the set of non-deterministic variables $y$ such that there exists a directed path from $x$ to $y$ given by $x, y_1, \ldots, y_n, y$, with all intermediate variables $(y_1, \ldots, y_n)$ being deterministic. The* non-deterministic parents *of a node $x$, denoted $ndparents(x)$, are the set of non-deterministic variables $y$ such that there exists a directed path from $y$ to $x$ given by $y, y_1, \ldots, y_n, x$, with all intermediate variables $(y_1, \ldots, y_n)$ being deterministic. The* deterministic children *of a node $x$, denoted $detchildren(x)$, are the set of deterministic variables $y$ that are children of $x$. The* deterministic parents *of a node $x$, denoted $detparents(x)$, are the set of deterministic variables $y$ that are parents of $x$.*

For instance, in the model in Figure 23, the non-deterministic children of $w_3$ are $o_1$ and $o_2$. Deterministic nodes can be removed from a graph by rewriting the equations represented into the remaining variables of the graph. Because some graphical operations do not apply to deterministic nodes, this removal is often done implicitly within a theorem. This goes as follows:

**Lemma 6.1** *A chain graph $G$ with nodes $X$ has deterministic nodes $Y \subset X$. The chain graph $G'$ is created by adding to $G$ a directed arc from every node to its non-deterministic children, and by deleting the deterministic nodes $Y$. The graphs $G$ and $G'$ are equivalent probability models on the nodes $X - Y$.*

The general formula for differentiating Bayesian networks with plates and deterministic nodes is given below in Lemma 6.2. This is nothing more than the chain rule for differentiation, but it is important to notice the network structure of the computation. When partial derivatives are computed over networks, there are local and global partial derivatives that can be different. Consider the feed-forward network of Figure 23 again. On this figure, place an extra arc from $w_4$ to $m_2$. Now consider the partial derivative of $m_2$ with respect to $w_4$. The value of $m_2$ is influenced by $w_4$ directly, as the new arc shows, and indirectly via $h_2$. When computing a partial derivative involving indirect influences, we need to differentiate between the direct and indirect effects. Various notations are used for this (Werbos et al., 1992; Buntine & Weigend, 1994). Here the notation of a local versus global derivative is used. The *local partial derivative* is subscripted with an $l$, $\partial/\partial_l$ and represents the partial derivative computed at the node using only the direct influences—the parents. For the example of the partial derivative of $m_2$ with respect to $w_4$, the various local partial derivatives combine to produce the global partial derivative:

$$\frac{\partial m_2}{\partial w_4} = \frac{\partial m_2}{\partial_l w_4} + \frac{\partial m_2}{\partial_l h_2}\frac{\partial h_2}{\partial_l w_4} .$$

This is equivalent to:

$$\frac{\partial m_2}{\partial w_4} = \frac{\partial m_2}{\partial_l w_4} + \frac{\partial m_2}{\partial_l h_2}\frac{\partial h_2}{\partial_l w_4} + \frac{\partial m_2}{\partial_l h_1}\frac{\partial h_1}{\partial_l w_4} ,$$

since $\frac{\partial h_1}{\partial_l w_4} = 0$.

In general, the (global) partial derivative for an index variable $\theta_i$ is the sum of the

- local partial derivative at the node containing $\theta_i$,

- the partial derivatives for each child of $\theta_i$ that is also a non-deterministic child, and

- combinations of (global) partial derivatives for deterministic children found by backward or forward propagation of derivatives.

**Lemma 6.2** (Differentiation). *A model $M$ is represented by a Bayesian network $G$ with plates and deterministic nodes on variables $X$. Denote the known variables in $X$ by $K$ and the unknown variables by $U = X - K$. Let the conditional probability represented by the graph $G$ be $p(U|K, M)$. Let $\theta$ be some unknown variable in the graph, and let $1_{nd(\theta)}$ be 1 if $\theta$ is non-deterministic and 0 otherwise. If $\theta$ occurs inside a plate then let $i$ be some arbitrary valid index ($i \in indval(\theta)$), otherwise let $i$ be null. Then:*

$$
\begin{aligned}
\frac{\partial \log p(U|K, M)}{\partial \theta_i} \;=\;& 1_{nd(\theta_i)} \frac{\partial \log p(\theta_i | parents(\theta_i))}{\partial_l \theta_i} \\
& + \sum_{x \in ndchildren(\theta_i) \cap children(\theta_i)} \frac{\partial \log p(x|parents(x))}{\partial_l \theta_i} \\
& + \sum_{x \in ndchildren(\theta_i)} \sum_{y \in detparents(x), y \neq \theta_i} \frac{\partial \log p(x|parents(x))}{\partial_l y} \frac{\partial y}{\partial \theta_i} \; .
\end{aligned}
\tag{17}
$$

*Furthermore, if $Y \subset U$ is some subset of the unknown variables, then the partial derivative of the probability of $Y$ given the known variables, $p(Y|K, M)$ is an expected value of the above probabilities:*

$$
\frac{\partial \log p(Y|K, M)}{\partial \theta_i} \;=\; \mathcal{E}_{U-Y|Y,K,M} \left( \frac{\partial \log p(U|K, M)}{\partial \theta_i} \right) \; .
\tag{18}
$$

Equation (17) contains only one global partial derivative which is inside the double sum on the right side. This is the partial derivative $\partial y / \partial \theta_i$ and can be computed from its local island of determinism using the chain rule of differentiation, for instance, using forward propagation from $\theta_i$'s deterministic children, or backward propagation from $\theta_i$'s non-deterministic parents.

To apply the Differentiation Lemma on problems like feed-forward networks or unsupervised learning, the lemma needs to be extended to chain graphs. This means differentiating Markov networks as well as Bayesian networks, and handling the expected value in Equation (18). These extensions are explained below after first giving two examples.

As a first example, consider the feed-forward network problem of Figure 23. By treating the two output units in the feed-forward network as a single variable, a Cartesian product $(o_1, o_2)$, the above Differentiation Lemma can now be applied directly to the feed-forward network model of Figure 23. This uses the simplification given in Section 2.4 with Comment 2.1. Let $P$ be the joint probability for the feed-forward network model, given in Equation (16). The non-deterministic children of $w_4$ are the single chain component consisting of the two variables $o_1$ and $o_2$. Its parents are the set $\{m_1, m_2\}$. Consider the Differentiation Lemma. There are no children of $w_4$ that are also non-deterministic, so the middle sum in Equation (17) of the Differentiation Lemma is empty. Then the lemma yields, after expanding out the inner most sum:

$$
\frac{\partial \log P}{\partial w} \;=\; \frac{\partial \log p(w)}{\partial_l w_4} +
$$

$$\sum_{i=1}^{N} \left( \frac{\partial \log p(o_{1,i}, o_{2,i} | \Sigma, m_1, m_2)}{\partial_l m_1} \frac{\partial m_1}{\partial_l w_4} + \frac{\partial \log p(o_{1,i}, o_{2,i} | \Sigma, m_1, m_2)}{\partial_l m_2} \frac{\partial m_2}{\partial_l w_4} \right) \ ,$$

where $p(o_{1,i}, o_{2,i} | \Sigma, m_1, m_2)$ is the two-dimensional Gaussian, and $\frac{\partial m_i}{\partial w_4}$ is from the global derivative but evaluates to a local derivative.

As a second example, reconsider the simple unsupervised learning problem given in the introduction to Section 3. The likelihood for a single datum given the model parameters is a marginal of the form:

$$p(var_1 = 1, var_2 = 0, var_3 = 1 | \phi, \theta) \ = \ \sum_{c=1}^{10} \phi_c \, \theta_{1,c} \, (1 - \theta_{2,c}) \, \theta_{3,c} \ .$$

Taking the logarithm of the full case probability $p(class, var_1, var_2, var_3 | \phi, \theta)$ reveals the vectors of components $w$ and $t$ of the exponential distribution:

$\log p(class, var_1, var_2, var_3 | \phi, \theta)$
$$= \ \sum_{c=1}^{10} 1_{class=c} \ \log \phi_c + \sum_{j=1}^{3} \sum_{c=1}^{10} \left( 1_{class=c, var_j=true} \ \log \theta_{j,c} + 1_{class=c, var_j=false} \ \log(1 - \theta_{j,c}) \right) \ .$$

Notice that the normalizing constant $Z(\phi, \theta)$ is 1 in this case. Consider finding the partial derivative $\partial \log p(var_1, var_2, var_3 | \phi, \theta) / \partial \theta_{2,5}$. This is done for each case when differentiating the posterior or the likelihood of the unsupervised learning model. Applying Equation (18) to this yields:

$$\frac{\partial \log p(var_1, var_2, var_3 | \phi, \theta)}{\partial \theta_{2,5}}$$
$$= \ \sum_{d=1}^{10} \frac{\partial \log \theta_{2,d}}{\partial \theta_{2,5}} \mathcal{E}_{class=c|var_1,var_2,var_3,\phi,\theta} \left( 1_{class=d, var_2=true} \right)$$
$$+ \frac{\partial \log(1 - \theta_{2,d})}{\partial \theta_{2,5}} \mathcal{E}_{class=c|var_1,var_2,var_3,\phi,\theta} \left( 1_{class=d, var_2=true} \right)$$
$$= \ \frac{1}{\theta_{2,5}} 1_{var_2=true} \, p(class = 5 | var_1, var_2, var_3, \phi, \theta)$$
$$+ \frac{1}{1 - \theta_{2,5}} 1_{var_2=false} \, p(class = 5 | var_1, var_2, var_3, \phi, \theta) \ .$$

Notice that the derivative is computed by doing first-order inference to find $p(class = 5 | var_1, var_2, var_3, \phi, \theta)$, as noted by Russell, Binder, and Koller (1994). This property holds in general for exponential family models with missing or unknown variables. Derivatives are calculated by some first-order inference followed by a combination with derivatives of the $w$ functions. Consider the notation for the exponential family introduced previously in Definition 4.1, where the functional form is:

$$p(x | y, \theta, M) \ = \ \frac{h(x, y)}{Z(\theta)} \exp \left( \sum_{i=1}^{k} w_i(\theta) t_i(x, y) \right) \ .$$

Consider the partial derivative of a marginal of this probability, $p(x - u | y, \theta, M)$, for $u \subset x$. Using Equation (18) in the Differentiation Lemma, the partial derivative becomes:

$$\frac{\partial \log p(x - u | y, \theta, M)}{\partial \theta} = \sum_{i=1}^{k} \frac{\partial w_i(\theta)}{\partial \theta} \mathcal{E}_{u|x-u,y,\theta} \left( t_i(x, y) \right) - \frac{\partial Z(\theta)}{\partial \theta} . \quad (19)$$

If the partition function is not known in closed form (the case with the Boltzmann machine) then the final derivative $\partial Z(\theta) / \partial \theta$ is approximated (the key formula for doing this is Equation (28) in Appendix B).

To extend the Differentiation Lemma to chain graphs, use the trick illustrated with the feed-forward network. First, interpret the chain graph as a Bayesian network on chain components, as done in Equation (7), then apply the Differentiation Lemma. Finally, evaluate necessary local partial derivatives with respect to $\theta_i$ of each individual chain component. Since undirected graphs are not necessarily normalized, this may present a problem. In general, there is an undirected graph $G'$ on variables $X \cup Y$. Following Theorem 2.1, the general form is:

$$p(X | Y) = \frac{\prod_{C \in Cliques(G')} f_C(C)}{\sum_X \prod_{C \in Cliques(G')} f_C(C)} .$$

The local partial derivative with respect to $x$ becomes:

$$\frac{\partial \log p(X | Y)}{\partial_l x} = \left( \sum_{C \in Cliques(G'), x \in C} \frac{\partial \log f_C(C)}{\partial_l x} \right) - \mathcal{E}_{X|Y} \left( \sum_{C \in Cliques(G'), x \in C} \frac{\partial \log f_C(C)}{\partial_l x} \right) . \quad (20)$$

The difficulty here is computing the expected value in the formula, which comes from the normalizing constant. Indeed, this computation forms the core of the early Boltzmann machine algorithm (Hertz et al., 1991). In general, this must be done using something like Gibbs sampling and the techniques of Section 7.1 can be applied directly.

## 6.2 Decomposing learning problems

Learning problems can be decomposed into sub-problems in some cases. While the material in this section applies generally to these sorts of decompositions, this section considers one simple example and then proves some general results on problems decomposition. Problem decompositions can also be recomputed on the fly to create a search through a space of models that takes advantage of decompositions that exist. A general result is also presented on incremental decomposition. These results are simple applications of known methods for testing independence (Frydenberg, 1990; Lauritzen et al., 1990), with some added complication because of the use of plates.

Consider the simple learning problem given in Section 3, Figure 11 over two multinomial variables $var_1$ and $var_2$, and two Gaussian variables $x_1$ and $x_2$. For this problem we have specified two alternative models, model $M_1$ and model $M_2$. Model $M_2$ has an additional arc going from the discrete variable $var_2$ to the real valued variable $x_1$. We will use this subsequently to discuss local search of these models evaluated by their evidence.

A manipulation of the conditional distribution for this model, making use of Lemma 2.1, yields, for model $M_1$, the conditional distribution given in Figure 24. When parameters, $\theta_1$, $\theta_2$ are *a priori* independent, and their data likelihoods do not introduce cross terms between them, the parameters become *a posteriori* independent as well. This occurs for $\theta_1$, $\theta_2$, and the set $\{\mu_1, \sigma_1\}$. This model simplification also implies the evidence for model $M_1$ decomposes similarly. Denote the sample of the variable $x_1$ as $x_{1,*} = x_{1,1}, \ldots, x_{1,N}$, and likewise for $var_1$ and $var_2$. In this case, the result is:

$$evidence(M_1) = p(var_{1,*}|M_1) \, p(var_{2,*}|var_{1,*}, M_1) \, p(x_{1,*}|var_{1,*}, M_1) \, p(x_{2,*}|x_{1,*}, var_{1,*}, M_1) \, . \tag{21}$$

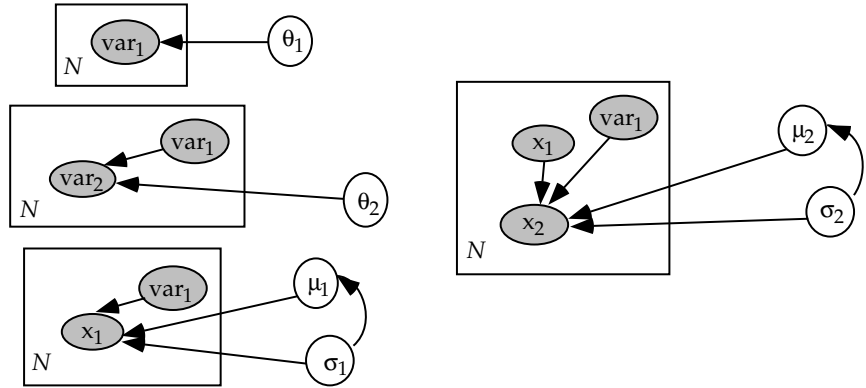

Figure 24: A simplification of model $M_1$

The evidence for model $M_2$ is similar except that the posterior distribution of $\mu_1$ and $\sigma_1$ is replaced by the posterior distribution for $\mu_1'$ and $\sigma_1'$.

This result is general, and applies to Bayesian networks, undirected graphs, and more generally to chain graphs. Similar results are covered by Dawid and Lauritzen (1993) for a family of models they call hyper-Markov. The general result described above is an application of the rules of independence applied to plates. This uses the notion of non-deterministic children and parents introduced in Definition 6.1. It also requires a notion of local dependence, which is called the Markov blanket, following Pearl (1988), since it is a generalization of the equivalent set for Bayesian networks.

**Definition 6.2** *We have a chain graph $G$ without plates. The* Markov blanket *of a node $u$ is all neighbors, non-deterministic parents, non-deterministic children, and non-deterministic parents of the children and their chain components:*

$$Markov\text{-}blanket(u) = neighbors(u) \cup ndparents(u) \cup ndchildren(u) \tag{22}$$
$$\cup \, ndparents(chain\text{-}components(ndchildren(u))) \, .$$

From Frydenberg (1990) it follows that $u$ is independent of the other non-deterministic variables in the graph $G$ given the Markov blanket.

To perform the simplification depicted in Figure 24, it is sufficient then to find the finest partitioning of the model parameters such that they are independent. The decomposition in Figure 24 represents the finest such partition of model $M_1$. The evidence for the model will then factor according to the partition, as given for model $M_1$ in Equation (21). For this task there is the following theorem, depicted graphically in Figure 25.

**Theorem 6.1** (Decomposition). *A model $M$ is represented by a chain graph $G$ with plates. Let the variables in the graph be $X$. There are $P$ possibly empty subsets of the variables $X$, $X_i$ for $i = 1, \ldots, P$ such that $unknown(X_i)$ is a partition of $unknown(X)$. This induces a decomposition of the graph $G$ into $P$ subgraphs $G_i$ where:*

- *the graph $G_i$ contains the nodes $X_i$ and any arcs and plates occurring on these nodes, and*

- *the potential functions for cliques in $G_i$ are equivalent to those in $G$.*

*The induced decomposition represents the unique finest equivalent independence model to the original graph if and only if $X_i$ for $i = 1, \ldots, P$ is the finest collection of sets such that, when ignoring plates, for every unknown node $u$ in $X_i$, its Markov blanket is also in $X_i$. This finest decomposition takes $O(|X|^2)$ to compute. Furthermore, the evidence for $M$ now becomes a product over each subgraph:*

$$evidence(M) \;=\; p(known(X_*)|M) \;=\; f_0 \prod_i f_i(known(X_{i,*})) \tag{23}$$

*for some functions $f_i$ (given in the proof).*

Figure 25 shows how this decomposition works when there are unknown nodes. Figure 25(a) shows the basic problem and Figure 25(b) shows the finest decomposition. Notice the bottom component cannot be further decomposed because the variable $x_1$ is unknown.

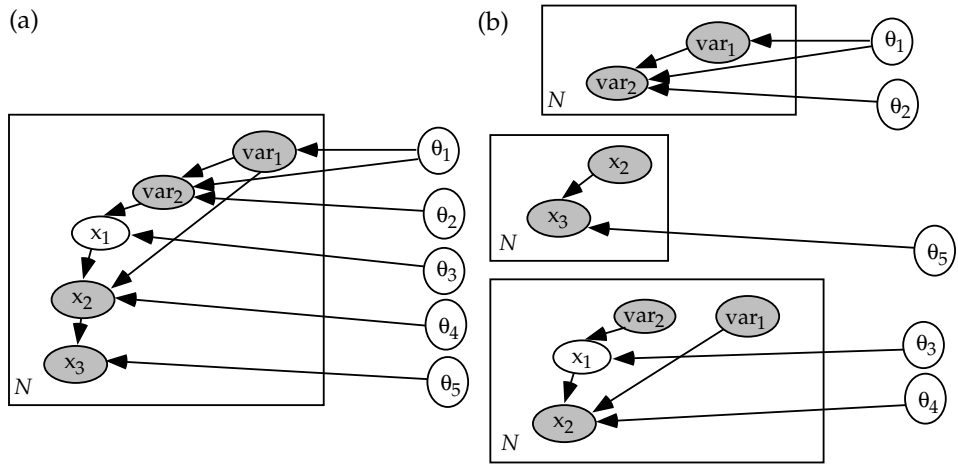

Figure 25: The incremental decomposition of a model

In some cases, the functions $f_i$ given in the Decomposition Theorem in Equation (23) have a clean interpretation: they are equal to the evidence for the subgraphs. This result can be obtained from the following corollary.

**Corollary 6.1.1** (Local Evidence). *In the context of Theorem 6.1, suppose there exists a set of chain components $\tau_j$ from the graph ignoring plates such that $X_j = \tau_j \cup ndparents(\tau_j)$, where $unknown(ndparents(\tau_j)) = \emptyset$. Then*

$$f_j(known(X_{j,*})) = p(known(\tau_j)_* | ndparents(\tau_j)_*, M) .$$

If we denote the $j$-th subgraph by model $M_j^S$, then this term is the conditional evidence for model $M_j^S$ given $ndparents(\tau_j)_*$. Denote by $M_0^S$ the maximal subgraph on known variables only (induced by $cliques_0$ as given in the proof of the Decomposition Theorem). If the condition of Corollary 6.1.1 holds for $M_j^S$ for $j = 0, 1, \ldots, P$, then it follows that the evidence for the model $M$ is equal to the product of the evidence for each subgraph:

$$evidence(M) = \prod_{i=0}^{P} evidence(M_i^S) . \tag{24}$$

This holds in general if the original graph $G$ is a Bayesian network, as used in learning Bayesian networks (Buntine, 1991c; Cooper & Herskovits, 1992).

**Corollary 6.1.2** *Equation (24) holds if the parent graph $G$ is a Bayesian network with plates.*

In general, we might consider searching through a family of graphical models. To do this local search (Johnson, Papdimitriou, & Yannakakis, 1985) or numerical optimization can be used to find high posterior models, or Markov chain Monte Carlo methods to select a sample of representative models, as discussed in Section 7.2. To do this, how to represent a family of models must be shown. Figure 26, for instance, is similar to the models of Figure 11 except that some arcs are hatched. This is used to indicate that these arcs are optional.

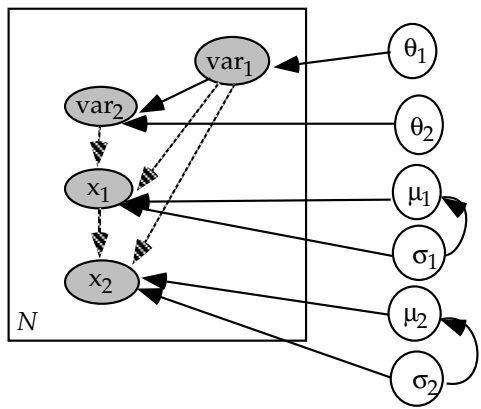

Figure 26: A family of models (optional arcs hatched)

To instantiate a hatched arc they can either be removed or replaced with a full arc. This graphical model then represents many different models, for all $2^4$ possible instantiations of the arcs. Prior probabilities for these models could be generated using a scheme such as in (Buntine, 1991c, p54) or (Heckerman et al., 1994), where a prior probability is assigned by a domain expert for different parts of the model, arcs and parameters, and the prior for a full model found by multiplication. The family of models given by Figure 26 includes those of Figure 11 as instances. During search or sampling, an important property is the Bayes factor for the two models, $Bayes\text{-}factor(M_2, M_1)$, as described in Section 3.1. Because of the Decomposition Theorem and its corollary, the Bayes factor for $M_2$ versus $M_1$ can be found by looking at local Bayes factors. The difference between models $M_1$ and $M_2$ is the parent for the variable $x_1$,

$$Bayes\text{-}factor(M_2, M_1) \;=\; \frac{p(x_{1,*}|var_{1,*}, var_{2,*}, M_2)}{p(x_{1,*}|var_{1,*}, M_1)} \;.$$

That is, the Bayes factor can be computed from only considering the models involving $\mu_1, \sigma_1$ and $\mu_1', \sigma_1'$.

This incremental modification of evidence, Bayes factors, and finest decompositions is also general, and follows directly from the independence test. A similar property for undirected graphs is given in (Dawid & Lauritzen, 1993). This is developed below for the case of directed arcs and non-deterministic variables. Handling deterministic variables will require repeated application of these results, because several non-deterministic variables may be effected when adding a single arc between deterministic variables.

**Lemma 6.3** (Incremental decomposition). *For a graph $G$ in the context of Theorem 6.1, we have two non-deterministic variables $U$ and $V$ such that $U$ is given. Consider adding or removing a directed arc from $U$ to $V$. To update the finest decomposition of $G$, there is a unique subgraph containing the unknown variables in ndparents(chain-component($V$)). To this subgraph add or delete an arc from $U$ to $V$, and add or delete $U$ to the subgraph if required.*

Shaded non-deterministic parents can be added at will to nodes in a graph, and the finest decomposition remains unchanged except for a few additional arcs. The use of hatched arcs in these contexts causes no additional trouble to the decomposition process. That is, the finest decomposition for a graph with plates and hatched directed arcs is formed as if the arcs where unhatched directed arcs. The evidence is adjusted during the search by adding the different parents as required.

## 6.3 Bayes factors for the exponential family

To make use of the decomposition results in a learning system it is necessary to be able to generate Bayes factors or evidence for the component models. For models in the exponential family, whose normalization constant is known in closed form, this turns out to be easy. If these exact computations are not available, various approximation methods can be used to compute the evidence or Bayes factors (Kass & Raftery, 1993); some are discussed in Section 7.3.

If the conjugate distribution for an exponential family model and its derivatives can be readily computed, then the Bayes factor for the model can be found in closed form. Along with the above decomposition methods, this result is an important basis of many fast Bayesian algorithms considering multiple models. It is used explicitly or implicitly in all Bayesian methods for learning decision trees, directed graphical models (with discrete or Gaussian variables), and linear regression (Buntine, 1991b; Spiegelhalter et al., 1993).

For instance, if the normalizing constant $Z_\theta(\tau)$ in Lemma 4.1 was known in closed form, then the Bayes factor can be readily computed.

**Lemma 6.4** *Consider the context of Lemma 4.1. Then the* model likelihood *or* evidence, *given by* $evidence(M) = p(x_1, \ldots, x_N | y_1, \ldots, y_N, M)$, *can be computed as:*

$$
\begin{aligned}
evidence(M) &= \frac{p(\theta|\tau)\prod_{j=1}^{N} p(x_j|y_j,\theta)}{p(\theta|\tau')} \\
&= \frac{Z_\theta(\tau')}{Z_\theta(\tau)Z_2^N} .
\end{aligned}
$$

For $y|x \sim$ Gaussian this involves multiplying out the two sets of normalizing constants for the Gaussian and Gamma distributions. The evidence for some common exponential family distributions is given in Appendix B in Table 5

For instance, consider the learning problem given in Figure 24. Assume that the variables $var_1$ and $var_2$ are both binary (0 or 1) and that the parameters $\theta_1$ and $\theta_2$ are interpreted as follows:

$$
\begin{aligned}
p(var_1 = 0 | \theta_1) &= \theta_1 , \\
p(var_2 = 0 | var_1 = 0, \theta_2) &= \theta_{2,0|0} , \\
p(var_2 = 0 | var_1 = 1, \theta_2) &= \theta_{2,0|1} .
\end{aligned}
$$

If we use Dirichlet priors for these parameters, as shown in Table 3, then the priors are:

$$
\begin{aligned}
(\theta_1, 1 - \theta_1) &\sim \text{Dirichlet}(\alpha_{1,0}, \alpha_{1,1}) , \\
(\theta_{2,0|j}, 1 - \theta_{2,0|j}) &\sim \text{Dirichlet}(\alpha_{2,0|j}, \alpha_{2,1|j}) \qquad \text{for } j = 0, 1 ,
\end{aligned}
$$

where $\theta_{2,0|0}$ is *a priori* independent of $\theta_{2,0|1}$. The choice of priors for these distributions is discussed in (Box & Tiao, 1973; Bernardo & Smith, 1994). Denote the corresponding sufficient statistics as $n_{1,j}$ (equal to the number of data where $var_1 = j$) and $n_{2,j|i}$ (equal to the number of data where $var_2 = j$ and $var_1 = i$). Then the first two terms of the evidence for model $M_1$, read directly from Table 5, can be written as:

$$
\begin{aligned}
p(var_{1,*}|M_1) &= \frac{Beta(n_{1,0} + \alpha_{1,0}, n_{1,1} + \alpha_{1,1})}{Beta(\alpha_{1,0}, \alpha_{1,1})} , \\
p(var_{2,*}|var_{1,*}, M_1) &= \frac{Beta(n_{2,0|0} + \alpha_{2,0|0}, n_{2,1|0} + \alpha_{2,1|0})}{Beta(\alpha_{2,0|0}, \alpha_{2,1|0})} \frac{Beta(n_{2,0|1} + \alpha_{2,0|1}, n_{2,1|1} + \alpha_{2,1|1})}{Beta(\alpha_{2,0|1}, \alpha_{2,1|1})} .
\end{aligned}
$$

Assume the variables $x_1$ and $x_2$ are Gaussian with means given by

$$
\begin{aligned}
\mu_{1|0} &\qquad \text{when } var_1 = 0, \\
\mu_{1|1} &\qquad \text{when } var_1 = 1, \\
\mu_{2|0,1} + \mu_{2|0,2}x_1 &\qquad \text{when } var_1 = 0 , \\
\mu_{2|1,1} + \mu_{2|1,2}x_1 &\qquad \text{when } var_1 = 1
\end{aligned}
$$

and variances $\sigma_{1|j}$ and $\sigma_{2|j}$ respectively. In this case, we split the data set into two parts, those when $var_1 = 0$, and those when $var_1 = 1$. Each get their own parameters, sufficient statistics, and contribution to the evidence. Conjugate priors from Table 3 in Appendix B (using $y|x \sim$ Gaussian) are indexed accordingly as:

$$\mu_{i|j}|\sigma_{i|j} \quad \sim \quad \text{Gaussian}(\mu_{0,i|j}, \frac{\Sigma_{0,i|j}}{\sigma_{i|j}^2}) \qquad \text{for } i = 1, 2 \text{ and } j = 0, 1 \ ,$$

$$\sigma_{i|j}^{-2} \quad \sim \quad \text{Gamma}(\delta_{0,i|j}/2, \beta_{0,i|j}) \qquad \text{for } i = 1, 2 \text{ and } j = 0, 1 \ .$$

Notice that $\Sigma_{0,i|j}$ is one-dimensional when $i = 0$ and two-dimensional when $i = 2$. Suitable sufficient statistics for this situation are read from Table 4 by looking at the data summaries used there. This can be simplified for $x_1$ because $d = 1$ and $y_1$ for the Gaussian is uniformly 1. Thus the sufficient statistics for $x_1$ become the means and variances for the different values of $var_1$. Denote $\overline{x_{1|0}}$ and $\overline{x_{1|1}}$ as the sample means of $x_1$ when $var_1 = 0, 1$, respectively, and $s_{1|0}^2$ and $s_{1|1}^2$ their corresponding sample variances. This cannot be done for the second case, so we use the notation from Table 4, where $\Sigma, \overline{\mu}, \beta$ from Table 4 become, respectively, $\Sigma_{2|j}, \overline{\mu_{2|j}}, \beta_{2|j}$. Change the vector $y$ to $(1, x_1)$ when making the calculations indicated here. The sufficient statistics are, for each case of $var_1 = j$:

$$S_{2|j} \quad = \quad \sum_{i=1}^{N} 1_{var_{1,i}=j} \, y_i y_i^\dagger \ ,$$

$$m_{2|j} \quad = \quad \sum_{i=1}^{N} 1_{var_{1,i}=j} \, x_i y_i \ ,$$

$$s_{2|j}^2 \quad = \quad \sum_{i=1}^{N} 1_{var_{1,i}=j} \, (x_i - \overline{\mu_{2|j}}^\dagger y_i)^2 \ .$$

The evidence for the last two terms can now be read from Table 5. This becomes:

$$p(x_{1,*}|var_{1,*}, M_1) \quad = \quad \prod_{j=0,1} \frac{\sqrt{\Sigma_{0,1|j}}}{\pi^{n_{1,j}/2} \sqrt{\Sigma_{0,1|j} + n_{1,j}}} \frac{\Gamma((\delta_{0,1|j} + n_{1,0})/2)}{\Gamma(\delta_{0,1|j}/2)\beta_{0,1|j}^{\delta_{0,1|j}/2}}$$
$$\left(\beta_{0,1|j} + s_{1|j}^2 + \frac{\Sigma_0 N}{\Sigma_0 + N}(\overline{x} - \mu_{0,1|j})^2\right)^{(\delta_{0,1|j}+n_{1,j})/2} \ ,$$

$$p(x_{2,*}|x_{1,*}, var_{1,*}, M_1) \quad = \quad \prod_{j=0,1} \frac{\det^{1/2}\Sigma_{0,2|j}}{\pi^{n_{1,j}/2}\det^{1/2}\Sigma_{2|j}} \frac{\Gamma((\delta_{0,2|j} + n_{1,0})/2)\beta_{2|j}^{(\delta_{0,2|j}+n_{1,j})/2}}{\Gamma(\delta_{0,2|j}/2)\beta_{0,2|j}^{\delta_{0,2|j}/2}} \ .$$

The final simplification of the model is given in Figure 27.

## 7. Approximate methods on graphical models

Exact algorithms for learning of any reasonable size invariably involve the recursive arc reversal theorem of Section 4.2. Most learning methods, however, use approximate algorithms at some level. The most common uses of the exponential family within approximation algorithms were summed up in Figure 22. Various other methods for inference on plates can be applied either at the model level or the parameter level: Gibbs sampling, first described in Section 7.1, other more general Markov chain Monte Carlo algorithms, EM style algorithms (Dempster, Laird, & Rubin, 1977), and various closed form approximations such as the mean field approximation, and the Laplace approximation (Berger, 1985; Azevedo-Filho & Shachter, 1994). This section summarizes the main families of these approximate methods.

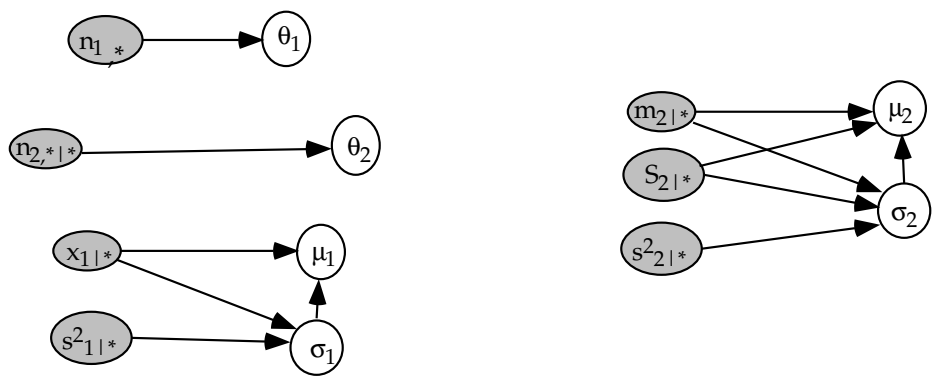

Figure 27: The full simplification of model $M_1$

## 7.1 Gibbs sampling

Gibbs sampling is the basic tool of simulation and can be applied to most probability distributions (Geman & Geman, 1984; Gilks et al., 1993a; Ripley, 1987) as long as the full joint has no zeros (all variable instantiations are possible). It is a special case of the general Markov chain Monte Carlo methods for approximate inference (Ripley, 1987; Neal, 1993). Gibbs sampling can be applied to virtually any graphical model whether there are plates, undirected or directed arcs, and whether the variables are real or discrete. Gibbs sampling does not apply to graphs with deterministic nodes, however, since these put zeroes in the full joint. This section describes Gibbs sampling without plates, as a precursor to discussing Gibbs sampling with plates in Section 7.2. On challenging problems, other forms of Markov chain Monte Carlo sampling can and should be tried. The literature is extensive.

Gibbs sampling corresponds to a probabilistic version of gradient ascent, although their goals of averaging as opposed to maximizing are fundamentally different. Gradient ascent in real valued problems corresponds to simple methods from function optimization (Gill et al., 1981) and in discrete problems corresponds to local repair or local search (Johnson et al., 1985; Minton, Johnson, Philips, & Laird, 1990; Selman, Levesque, & Mitchell, 1992). Gibbs sampling varies gradient ascent by introducing a random component. The algorithm usually tries to ascend, but will sometimes descend, as a strategy for exploring further around the search space. So the algorithm tends to wander around local maxima with occasional excursions to other regions of the space. Gibbs sampling is also the core algorithm of simulated annealing if temperature is held equal to one (van Laarhoven & Aarts, 1987).

To sample a set of variables $X$ according to some non-zero distribution $p(X)$, initialize $X$ to some value and then repeatedly resample each variable $x \in X$ according to its conditional probability $p(x|X - \{x\})$. For the simple medical problem of Figure 2, suppose the value of symptoms is known, and the remaining variables are to be sampled, then do as follows:

1. Initialize the remaining variables somehow.

2. Repeat the following for $i = 1, 2, 3, \ldots$, and record the sample of $Age_i, Occ_i, Clim_i, Dis_i$ at the end of each cycle.

   (a) Reassign $Age$ by sampling it according to the conditional:

   $$p(Age|Occ, Clim, Dis, Symp) .$$

   That is, take the values of $Occ, Clim, Dis, Symp$ as given and compute the resulting conditional distribution on $Age$. Then sample $Age$ according to that distribution.

   (b) Reassign $Occ$ by sampling it according to the conditional:

   $$p(Occ|Age, Clim, Dis, Symp) .$$

   (c) Reassign $Clim$ by sampling it according to the conditional:

   $$p(Clim|Age, Occ, Dis, Symp) .$$

   (d) Reassign $Dis$ by sampling it according to the conditional:

   $$p(Dis|Age, Clim, Occ, Symp) .$$

This sequence of steps is depicted in Figure 28. In this figure, the basic graph has been re-arranged for each step to represent the dependencies that arise during the sampling process. This uses the arc reversal and conditioning operators introduced previously.

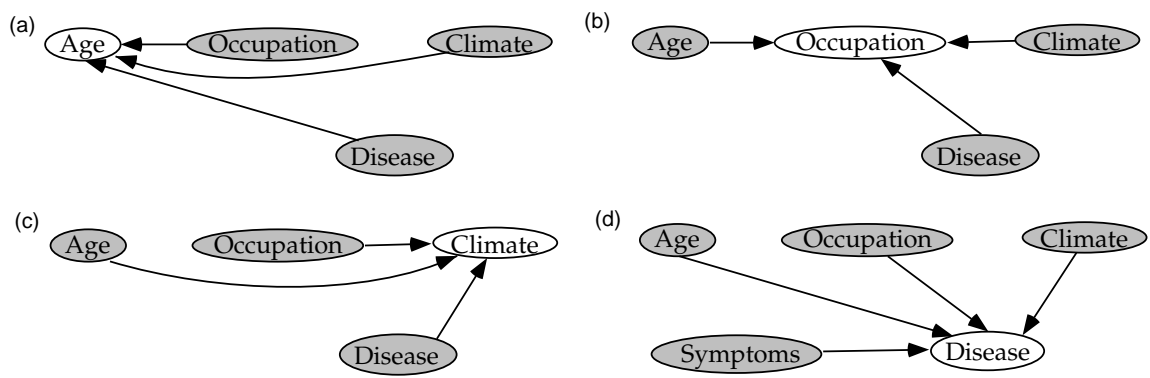

Figure 28: Gibbs sampling on the medical example

The effect of sampling is not immediate. $Age_2, Occ_2, Clim_2, Dis_2$ is conditionally dependent on $Age_1, Occ_1, Clim_1, Dis_1$, and in general so is $Age_i, Occ_i, Clim_i, Dis_i$ for any $i$. However the effect of the sampling scheme is that in the long run, for large $i$, $Age_i, Occ_i, Clim_i, Dis_i$ is approximately generated according to $p(Age, Occ, Clim, Dis | Symp)$ independently of $Age_1, Occ_1, Clim_1, Dis_1$. In Gibbs sampling, all the conditional sampling is done in accordance with the original distribution, and since this is a stationary process, in the long run the samples converge to the stationary distribution or fixed-point of the process. Methods for making subsequent samples independent are known as regenerative simulation (Ripley, 1987) and correspond to sending the temperature back to zero occasionally.

With this sample different quantities such as the probability a patient will have $Age > 20$ and $Clim = tropical$ given $Symp$ can be estimated. This is done by looking at the frequency of this event in the generated sample. The justification for this is the subject of Markov process theory (Çinlar, 1975, Theorem 2.26). The following result, presented informally, applies:

**Comment 7.1** *Let $x_1, x_2, \ldots, x_I$ be a sequence of discrete variables from a Gibbs sampler for the distribution $p(x) > 0$. Then the average of $g(x_i)$ approaches the expected value with probability 1 as $I$ approaches infinity:*

$$\frac{1}{I} \sum_{i=1}^{I} g(x_i) \;\longrightarrow\; \overline{g(x)} = \mathcal{E}\left(g(x)\right) \ .$$

*Further, for a second function $h(x_i)$, the ratio of two sample averages for $g$ and $h$ approaches their "true" ratio:*

$$\frac{\sum_{i=1}^{I} g(x_i)}{\sum_{i=1}^{I} h(x_i)} \;\longrightarrow\; \frac{\overline{g(x)}}{\overline{h(x)}} \ .$$

*This is used to approximate conditional expected values.*

To complete this procedure it is necessary to know how many Gibbs samples to take, how large to make $I$, and how to estimate the error in the estimate. Both these questions have no easy answer but heuristic strategies exist (Ripley, 1987; Neal, 1993).

For Bayesian networks this scheme is easy in general since the only requirement when sampling from $p(x | X - \{x\})$ is the conditional distribution for nodes connected to $x$, and the global probabilities do not need to be calculated. Notice, for instance, that in Figure 28 some sampling operations do not require all five variables. The general form for Bayesian networks given in Equation (2) goes as follows:

$$p(x | X - \{x\}) \;=\; \frac{p(X)}{\sum_x p(X)} \;=\; \frac{p(x | parents(x)) \prod_{y \,:\, x \in parents(y)} p(y | parents(y))}{\sum_x p(x | parents(x)) \prod_{y \,:\, x \in parents(y)} p(y | parents(y))} \ .$$

Notice the product is over a subset of variables. Only include conditional distributions for variables that have $x$ as a parent. Thus, the formula only involves examining the parents, children and children's parents of $x$, the so-called Markov blanket (Pearl, 1988). Also, notice normalization is only required over the single dimension changed in the current cycle, done in the denominator. For $x$ discrete, these conditional probabilities can be enumerated and direct sampling done for $x$.

The kind of simplification above for Bayesian networks also applies to undirected graphs and chain graphs, with or without plates. Here, modify Equation (4):

$$p(x | X - \{x\}) \;=\; \frac{\exp\left(\sum_{C \,:\, x \in C \in cliques(G)} f_C(C)\right)}{\sum_x \exp\left(\sum_{C \,:\, x \in C \in cliques(G)} f_C(C)\right)} \ . \tag{25}$$

In this formula, ignore all cliques not containing $x$ so, again, Gibbs sampling only computes with information local to the node. Also, the troublesome normalization constant does not have to be computed because the probability is a ratio of functions and so cancels out. As before, normalization is only required over the single dimension $x$.

## 7.2 Gibbs sampling on plates

Many learning problems can be represented as Bayesian networks. For instance, the simple unsupervised learning problem represented in Figure 10 is a Bayesian network once the plate is expanded out. It follows that Gibbs sampling is readily applied to learning as a general inference algorithm (Gilks et al., 1993a, 1993b).

Consider a simplified example of this unsupervised learning problem. In this model, assume that each variable $var_1$ and $var_2$ belongs to a mixture of Gaussians of known variance equal to 1.0. This simple model is given in Figure 29. For a given class, $class = c$, the variables $var_1$ and $var_2$ are distributed as Gaussian with means $\mu_{1,c}$ and $\mu_{2,c}$. In the uniform, unit-variance case the distribution for each sample is given by:

$$p(var_1, var_2 | , \phi, \mu, M) = \sum_c \phi_c N(var_1 - \mu_{1,c}) N(var_2 - \mu_{2,c}) ,$$

where $N( , )$ is the one-dimensional Gaussian probability density function with standard deviation of 1. This model might seem trivial, but if the standard deviation were to vary as well, the model corresponds to a Kernel density estimate so can approximate any other distribution arbitrarily well using sufficient number of tiny Gaussians.

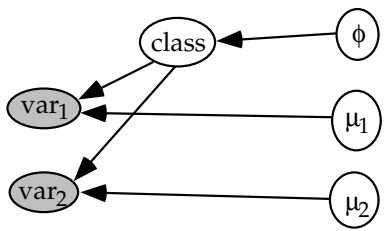

Figure 29: Unsupervised learning in two dimensions

In this simplified Gaussian mixture model, the sequence of steps for Gibbs sampling goes as follows:

1. Initialize the variables $\phi_c, \mu_{1,c}, \mu_{2,c}$ for each class $c$.

2. Repeat the following and record the sample of $\phi_c, \mu_{1,c}, \mu_{2,c}$ for each class $c$ at the end of each cycle.

   (a) For $i = 1, \ldots, N$, reassign $class_i$ according to the conditional:

   $$p(class_i \mid var_{1,i}, var_{2,i}, \phi, \mu_1, \mu_2) .$$

   (b) Reassign the vector $\phi$ by sampling according to the conditional:

   $$p(\phi | class_i \; : \; i = 1, \ldots, N) .$$

(c) Reassign the vector $\mu_1$ (and $\mu_2$) by sampling according to the conditional:

$$p(\mu_1 | var_{1,i}, class_i \; : \; i = 1, \ldots, N) \, .$$

Figure 30(a) illustrates Step 2(a) in the language of graphs. Figure 30(b) illustrates

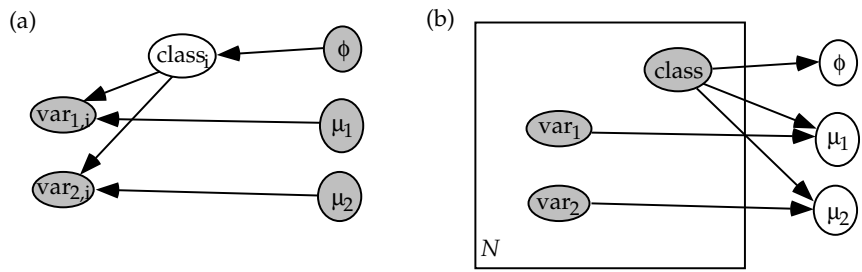

Figure 30: Gibbs sampling in the unsupervised learning problem

Steps 2(b) and 2(c). Step 2(a) represents the standard sampling operation using inference on Bayesian networks without plates. Step 2(b) and 2(c) are also easy to perform because in this case the distributions are exponential family, and the graph matches the conditions for Lemma 6.1.2. Therefore, each of the model parameters $\phi, \mu_1, \mu_2$ are *a posteriori* independent and their distribution is known in closed form, with the sufficient statistics calculated in $O(N)$ time.

One important caveat in the use of Gibbs sampling for learning is the problem of symmetry. In the above description, there is nothing to distinguish class 1 from class 2. Initially, the class centers $\mu$ for the above process will remain distinct. Asymptotically, since there is nothing in the problem definition to distinguish between class 1 and class 2, they will appear indistinguishable. This problem is handled by symmetry breaking: force $\mu_{1,c} < \mu_{2,c}$.

Gibbs sampling applies whenever there are variables associated with the data that are not given. Hidden or latent variables are an example. Incomplete data (or missing values) (Quinlan, 1989), robust methods and modeling of outliers, and various density estimation and non-parametric methods all fall in this family of models (Titterington et al., 1985). Gibbs sampling generalizes to virtually any graphical model with plates and unshaded nodes inside the plate; the sequence of sampling operations will be much the same as in Figure 30. If the underlying distribution is exponential family, for instance, Lemma 5.1 applies after shading all nodes inside the plate; each full cycle is guaranteed to be linear time in the sample size. The algorithm in the exponential family case is summed up in Figure 31. Figure 31(I) shows the general learning problem, extending the mixture model of Figure 22. This same structure appears in Figure 29. However, in Figure 31(I) the sufficient statistics $T(x_*, u_*)$ are also shown. Figure 31(II) shows more of the algorithm, generalizing Figure 30(a) and (b). Again the role of sufficient statistics is shown. The sampling in Figure 30(b) first computes the sufficient statistics and then sampling applies from that.

Thomas, Spiegelhalter, and Gilks (1992)(Gilks et al., 1993b) have taken advantage of this general applicability of sampling to create a compiler that converts a graphical representation of a data analysis problem, with plates, into a matching Gibbs sampling algorithm. This scheme applies to a broad variety of data analysis problems.

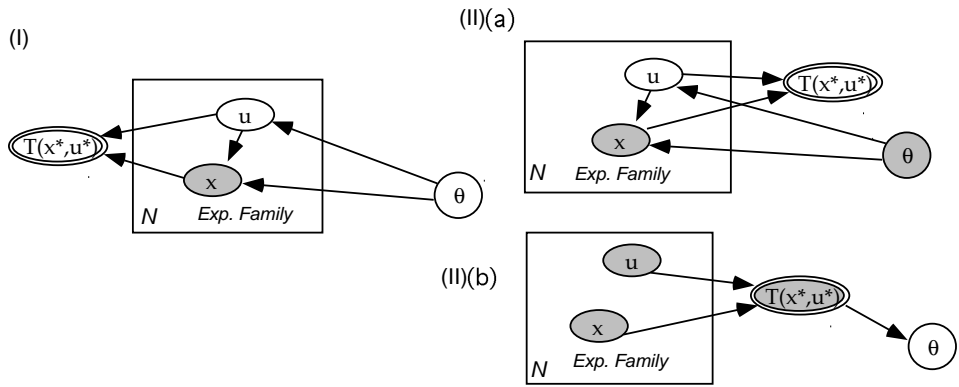

Figure 31: Gibbs sampling with the exponential family

## 7.3 A closed form approximation

What would happen to Gibbs sampling if the number of cases in the training sample, $N$, was large compared to the number of unknown parameters in the problem? A good way to think of this is: first, if $N$ is sufficiently large, then the samples of the model parameters $\phi$ and $\mu_1, \mu_2$ will tend to drift around their mean because their posterior variance would be $O(I/N)$. That is, after the $i$-th step, the sample is $\phi_i, \mu_{1,i}, \mu_{2,i}$. The $i+1$-th sample $\phi_{i+1}, \mu_{1,i+1}, \mu_{2,i+1}$ would be conditionally dependent on these, but because $N$ is large, the posterior variance of the $i+1$-th sample given the $i$-th sample would be small so that:

$$\phi_{i+1} \approx \mathcal{E}_{\phi_i, \mu_{1,i}, \mu_{2,i}} (\phi_{i+1}) \ .$$

This approximation is used with Markov chain Monte Carlo methods by the *mean field method* from statistical physics, popular in neural networks (Hertz et al., 1991). Rather than sampling a sequence of parameters $\theta, \theta_1, \theta_2, \ldots, \theta_i$, according to some scheme, use the deterministic update:

$$\overline{\theta_{i+1}} = \mathcal{E}_{\theta_i = \overline{\theta_i}} (\theta_{i+1}) \ , \tag{26}$$

where the expected value is according to the sampling distribution. This instead generates a deterministic sequence $\overline{\theta_1}, \overline{\theta_2}, \ldots, \overline{\theta_i}$ that under reasonable conditions converges to some maxima. This kind of approach leads naturally to the EM algorithm, which will be discussed in Section 7.4.

## 7.4 The expectation maximization (EM) algorithm

The expectation maximization algorithm, widely known as the EM algorithm, corresponds to a deterministic version of Gibbs sampling used to search for the MAP estimate for model parameters (Dempster et al., 1977). It is generally considered to be faster than gradient descent. Convergence is slow near a local maxima so some implementations switch to conjugate gradient or other methods (Meilijson, 1989) when near a solution. The computation used to find the derivative is similar to the computation used for the EM algorithm, so this does not require a great deal of additional code. Also, the determinism means the EM algorithm no longer generates unbiased posterior estimates of model parameters. The intended gain is speed, not accuracy. The EM algorithm can generally be applied to exponential family models wherever Gibbs sampling can be applied. The correspondence between EM and Gibbs is shown below.

Consider again the simple unsupervised learning problem represented in Figure 29 (Section 7.2). In this case, the sequence of steps for the EM algorithm is similar to that for the Gibbs sampler. The EM algorithm works on the means or modes of unknown variables instead of sampling them. Rather than sampling the set of classes and thereby computing sufficient statistics so that a distribution for $\phi$ and $\mu_1, \mu_2$ can be found, a sequence of class means are generated and thereby used to compute *expected sufficient statistics*. Likewise, instead of sampling new parameters $\phi$ and $\mu_1, \mu_2$, modes are then computed from the expected sufficient statistics.

Consider again the unsupervised learning problem in Figure 9. Suppose there are 10 classes and that the three variables $var_1, var_2, var_3$ are finite valued and discrete and modeled with a multinomial with probabilities conditional on the class value $class_i$. The sufficient statistics in this case are all counts: $n_j$ is the number of cases where the class is $j$; $n_{v,k|j}$ is the number of cases where $class = j$ and $var_v = k$:

$$
\begin{aligned}
n_j &= \sum_{i=1}^{N} 1_{class_i = j} \ , \\
n_{v,k|j} &= \sum_{i=1}^{N} 1_{class_i = j} 1_{var_{v,i} = k} \ .
\end{aligned}
$$

The *expected* sufficient statistics computed from the rules of probability for a given set of parameters $\phi$ and $\theta_1, \theta_2, \theta_3$ are given by:

$$
\begin{aligned}
\overline{n_j} &= \sum_{i=1}^{N} p(class_i = j | var_{1,i}, var_{2,i}, var_{3,i}, \phi, \theta_1, \theta_2, \theta_3) \ , \\
\overline{n_{v,k|j}} &= \sum_{i=1}^{N} p(class_i = j | var_{1,i}, var_{2,i}, var_{3,i}, \phi, \theta_1, \theta_2, \theta_3) \, 1_{var_{v,i} = k} \ .
\end{aligned}
$$

Thanks to Lemma 4.2, these kinds of expected sufficient statistics can be computed for most exponential family distributions. Once sufficient statistics are computed for any of the distributions posterior means or modes of the model parameters (in this case $\phi$ and $\theta_1, \theta_2, \theta_3$) can be found.

1. Initialize the parameters $\phi$ and $\theta_1, \theta_2, \theta_3$.

2. Repeat the following until some convergence criteria is met:

   (a) Compute the expected sufficient statistics $\overline{n_j}$ and $\overline{n_{v,k|j}}$.

   (b) Recompute $\phi$ and $\theta_1, \theta_2, \theta_3$ to be equal to their mode conditioned on the sufficient statistics. For many posterior distributions, these can be found in standard tables, and in most cases found via Lemma 4.2. For instance, using the mean for $\phi$ gives:

$$
\phi_j = \frac{\overline{n_j} + \alpha_j}{\sum_j (\overline{n_j} + \alpha_j)} \ .
$$

All of the other Gibbs sampling algorithms discussed in Section 7.2 can be similarly placed in this EM framework. When the mode is used in Step 2(b), ignoring numerical problems, the EM algorithm converges on a local maxima of the posterior distribution for the parameters (Dempster et al., 1977). The general method is summarized in the following comment (Dempster et al., 1977).

**Comment 7.2** *The conditions of Lemma 5.1 apply with data variables $X$ inside the plate and model parameters $\theta$ outside. In addition, some of the variables $U \subset X$ are latent, so they are unknown and unshaded. Some of the remaining variables are sometimes missing, so, for the data $X_i$, variables $V_i \subset (X - U)$ are not given. This means the data given for the $i$-th datum is $X - U - V_i$ for $i = 1, \ldots N$. The EM algorithm goes as follows:*

*E-step:  The contribution to the expected sufficient statistics for each datum is:*

$$ET_i \;=\; \mathcal{E}_{U_i, V_i \mid X - U_i - V_i, \theta} \left( t(X_i) \right) \; .$$

*The expected sufficient statistic is then $ET \;=\; \sum_{i=1}^N ET_i$.*

*M-step:  Maximize the conjugate posterior using the expected sufficient statistics $ET$ in place of the sufficient statistics using the MAP approach for this distribution.*

*The fixed point of this algorithm is a local maxima of the posterior for $\theta$.*

Figure 31(II) for Gibbs sampling in Section 7.2 illustrates the steps in EM nicely. Figure 31(II)(a) corresponds to the E-step. The expected sufficient statistics are found from the parameters $\theta$. Rather than sampling $u$, and therefore computing the sufficient statistics, the expected sufficient statistics are computed. Figure 31(II)(b) corresponds to the M-step. Here, given the sufficient statistics, the mean or mode of the parameters $\theta$ are computed instead of being sampled. EM is therefore Gibbs with a mean/mode approximation done at the two major sampling steps of the algorithm.

In some cases, the expected sufficient statistics can be computed in closed form. Assume the exponential family distribution for $p(X|\theta)$ has a known normalization constant $Z(\theta)$ and the link function $w^{-1}$ exists. For some $i$, the normalizing constant for the exponential family distribution $p(U_i, V_i | X - U_i - V_i, \theta)$ is known in closed form. Denote it by $Z_i(\theta)$. Then using the notation of Theorem 4.1:

$$ET_i \;=\; \left\{ \begin{array}{ll} t(X_i) & \text{if } U = V_i = \emptyset \; , \\[2mm] \dfrac{\mathrm{d}\, w(\theta)}{\mathrm{d}\theta}^{-1} \dfrac{\mathrm{d} \log Z_i(\theta)}{\mathrm{d}\theta} & \text{otherwise} \; . \end{array} \right.$$

# 8. Partial exponential models

In some cases, only an initial, inner part of a learning problem can be handled using the recursive arc reversal theorem of Section 4.3. In this case, simplify what can be simplified, and then solve the remainder of the problem using a generic method like the MAP approximation. This section presents several examples: linear regression with heterogeneous variance, feed-forward networks with a linear output layer, and Bayesian networks.

This general process was depicted in the graphical model for the partial exponential family of Figure 22. This is an abstraction used to represent the general process. Consider the problem of learning a Bayesian network, both structure and parameters, where the distribution is exponential family given the network structure. The variable $T$ is a discrete variable indicating which graphical structure is chosen for the Bayesian network. The variable $X$ represents the full set of variables given for the problem. The variable $\Theta_T$ represents the distributional parameters for the Bayesian network, and is the part of the model that is conveniently exponential family. That is, $p(X|\Theta_T, T)$ will be treated as exponential for different $\Theta_T$ and hold $T$ fixed. The sufficient statistics in this case are given by $ss(X*, T)$. In this case, the subproblem that conveniently falls in the exponential family, $p(\Theta_T|X*, T)$ is simplified, but it is necessary to resort to the more general learning techniques of previous sections to solve the remaining part of the problem, $p(T|X*)$.

## 8.1 Linear regression with heterogeneous variance

Consider the heterogeneous variance problem given in Figure 32. This shows a graphical model for the linear regression problem of Section 4.4 modified to the situation where the standard deviation is heterogeneous, so it is a function of the inputs $x$ as well. In this case,

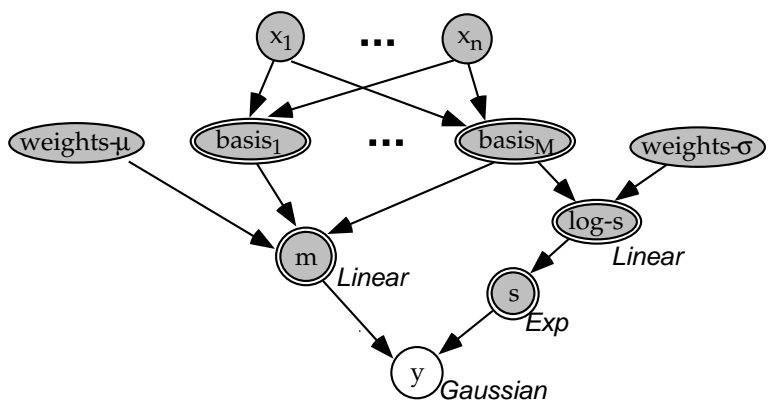

Figure 32: Linear regression with heterogeneous variance

the standard deviation $s$ is not given but is computed via:

$$s = \exp\left(\sum_{i=1}^{m} weights\text{-}\sigma_i \ basis_i(x)\right) .$$

The exponential transformation guarantees that the standard deviation $s$ will also be positive.

The corresponding learning model can be simplified to the graph in Figure 33. Compare this with the model given in Figure 21. What is the difference? In this case, the sufficient statistics exist, but they are shown to be deterministically dependent on the sample and ultimately on the unknown parameters for the standard deviation $weights\text{-}\sigma$. If the parameters for the standard deviation were known then the graph could be reduced to Figure 21. Computationally, this is an important gain. It says that for a given set of values for $weights\text{-}\sigma$, calculation can be done that is linear time in the sample size to arrive at a characterization of what the parameters from the mean, $weights\text{-}\mu$, should be. In short, one half of a problem, $p(weights\text{-}\mu|weights\text{-}\sigma, y_i, x_{.,i} : i = 1, \ldots, N)$, is well understood. To do a search for the MAP solution, search the space of parameters for the standard deviation, $weights\text{-}\sigma$, since the remaining ($weights\text{-}\mu$) is given.

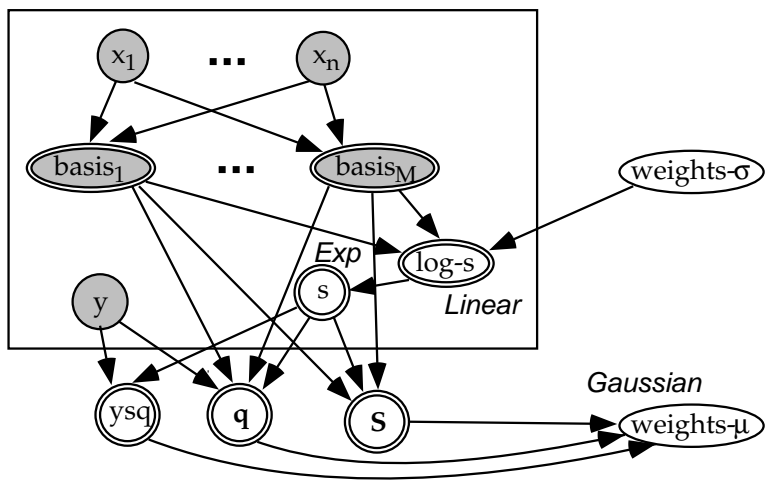

Figure 33: The heterogeneous variance problem with the plate simplified

Another variation of linear regression replaces the Gaussian error function with a more robust error function such as Student's $t$ distribution, or an $L_q$ norm for $1 < q < 2$. By introducing a convolution, these robust regression models can be handled by combining the EM algorithm with standard least squares (Lange & Sinsheimer, 1993).

## 8.2 Feed-forward networks with a linear output layer

A similar example is the standard feed-forward network where the final output layer is linear. This situation is given by Figure 23 if we change the deterministic functions for $m_1$ and $m_2$ to be linear instead of sigmoidal. In this case Lemma 5.1 identifies that when the weight vectors $w_3, w_4$, and $w_5$ are assumed given, the distribution is in the exponential family. Thus the simplification to Figure 34 is possible using the standard sufficient statistics for multivariate linear regression. Algorithmically, this implies that given values for the internal weight vectors $w_3, w_4$, and $w_5$, and assuming a conjugate prior holds for the output weight vectors $w_1$ and $w_2$, the posterior distribution for the output weight vectors $w_1$ and $w_2$, and their means and variances, can be found in closed form. The evidence for $w_1$ and $w_2$ given $w_3, w_4$ and $w_5$, $p(y_*|x_{1,*}, \ldots, x_{n,*}, w_3, w_4, w_5, M)$, can also be computed using the exact method of Lemma 6.4, so therefore the posterior for $w_3, w_4$, and $w_5$,

$$p(w_3, w_4, w_5|y_*, x_{1,*}, \ldots, x_{n,*}, M) \ \propto \ p(w_3, w_4, w_5|M) \, p(y_*|x_{1,*}, \ldots, x_{n,*}, w_3, w_4, w_5, M) \, ,$$

can be computed in closed form up to a constant. This effectively cuts the problem into two pieces—$w_3, w_4$ and $w_5$ followed by $w_1$ and $w_2$ given $w_3, w_4$, and $w_5$—and provides a clean solution to the second piece.

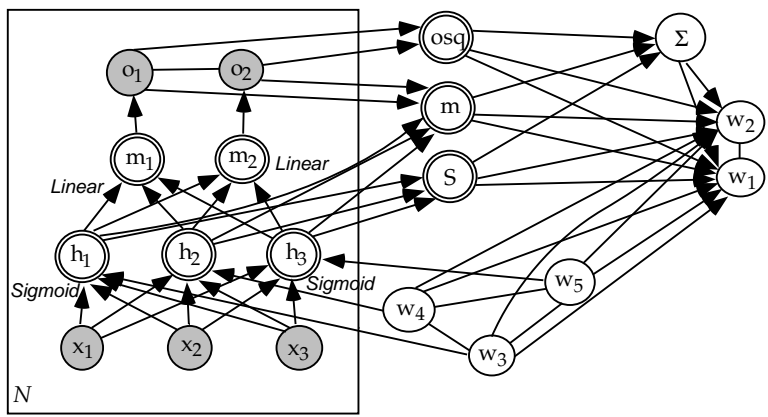

Figure 34: Simplified learning of a feed-forward network with linear output

## 8.3 Bayesian networks with missing variables

Class probability trees and discrete Bayesian networks can be learned efficiently by noticing that their basic form is exponential family (Buntine, 1991b, 1991a, 1991c; Cooper & Herskovits, 1992; Spiegelhalter et al., 1993). Take, for instance, the family of models specified by the Bayesian network given in Figure 26. In this case, the local evidence corollary, Corollary 6.1.1, applies. The evidence for Bayesian networks generated from this graph is therefore a product over the nodes in the Bayesian network. If we change a Bayesian network by adding or removing an arc, the Bayes factor is therefore simply the local Bayes factor for the node, as mentioned in the incremental decomposition lemma, Lemma 6.3. Local search is then quite fast, and Gibbs sampling over the space of Bayesian networks is possible. A similar situation exists with trees (Buntine, 1991b). The same results apply to any Bayesian network with exponential family distributions at each node, such as Gaussian or Poisson. Results for Gaussians are presented, for instance, in (Geiger & Heckerman, 1994).

This local search approach is a MAP approach because it searches for the network structure maximizing posterior probability. More accurate approximation can be done by generating a Markov chain of Bayesian networks from the search space of Bayesian networks. Because the Bayes factors are readily computed in this case, Gibbs sampling or Markov chain Monte Carlo schemes can be used. The scheme given below is the Metropolis algorithm (Ripley, 1987). This only looks at single neighbors until a successor is found. This is done be repeating the following steps:

1. For the initial Bayesian network $G$, randomly select a neighboring Bayesian network $G'$ differing only by an arc.

2. Compute *Bayes-factor*$(G', G)$ by making the decompositions described in Theorem 6.1, doing a local computation as described in Lemma 6.3, and using the Bayes factors computed with Lemma 6.4.

3. Accept the new Bayesian network $G'$ with probability given by:

$$\min\left(1, \textit{Bayes-factor}(G', G)\frac{p(G')}{p(G)}\right) \ .$$

If accepted, assign $G'$ to $G$, otherwise $G$ remains unchanged.

A local maxima Bayesian network could be found concurrently, however, this scheme generates a set of Bayesian networks appropriate for model averaging and expert evaluation of the space of potential Bayesian networks. Of course, initialization might search for local maxima to use as a reference. This sampling scheme was illustrated in the context of averaging in Figure 12.

This scheme is readily adapted to learn the structure and parameters of a Bayesian network with missing or latent variables. For the Metropolis algorithm, add Step 4, which resamples the missing data and latent variables.

4. For the current complete data and Bayesian network $G$, compute the predictive distribution for the missing data or latent variables. Use this to resample the missing data or latent variables to construct a new set of complete data (for subsequent use in computing Bayes factors).

## 9. Conclusion

The marriage of learning and graphical models presented here provides a framework for understanding learning. It also provides a framework for developing a learning or data analysis toolkit, or more ambitiously, a software generator for learning algorithms. Such a toolkit combines two important components: a language for representing a learning problem together with techniques for generating a matching algorithm. While a working toolbox is not demonstrated, a blueprint is provided to show how it could be constructed, and the construction of some well-known learning algorithms has been demonstrated. Table 1 lists some standard problems, the derivation of algorithms using the operations from the previous chapters, and where in the text they are considered. The notion of a learning toolkit is not new, and can be seen in the BUGS system by Thomas, Spiegelhalter, and Gilks (1992)(Gilks et al., 1993b), in the work of Cohen (1992) for inductive logic programming, and emerging in software for handling generalized linear models (McCullagh & Nelder, 1989; Becker, Chambers, & Wilks, 1988).

There is an important role for a data analysis toolkit. Every problem has its own quirks and requirements. Knowledge discovery, for instance, can vary in many ways depending on the user-defined notion of interestingness. Learning is often an embedded task in a larger system. So while there are some easy applications of learning, generally learning applications require special purpose development of learning systems or related support software. Sometimes, this can be achieved by patching together some existing techniques or by decomposing a problem into subproblems. Nevertheless, the decomposition and patching of learning algorithms with inference and decision making can be formalized and understood within graphical models. In some ways the S system plays the role of a toolkit (Chambers & Hastie, 1992). It provides a system for prototyping learning algorithms, includes the ability to handle generalized linear models, does automatic differentiation of expressions, and includes many statistical and mathematical functions useful as primitives. The language of graphical models is best viewed as an additional layer on top of this kind of system. Note, also, that it is impractical to assume that a software generator could create algorithms competitive with current finely tuned algorithms, for instance, for hidden Markov models. However, a software toolkit for learning could be used to prototype an algorithm that could later be refined by hand.

| Problem | Method | Sections |
|---|---|---|
| Bayesian networks and exponential family conditionals | Decomposition of exact Bayes factors, with local search or Gibbs sampling to generate alternative models | 6.2, 6.3, 8.3 |
| Bayesian networks with missing/latent variables, and other unsupervised learning models | Gibbs sampling or EM making use of above techniques where possible | 7.2, 7.4, 8.3 |
| Feed-forward networks | MAP method by exact computation of derivatives | 6.1 |
| Feed-forward networks with linear output | As above with an initial removal of the linear component | 8.2 |
| Linear regression and extensions | Least squares, EM, and MAP | 4.4, 8.1 |
| Generalized linear models | MAP method by exact computation of derivatives | 2.3, 6.1 |

Table 1: Derivation of learning algorithms

The combination of learning and graphical models shares some of the superior aspects of each of the different learning fields. Consider the philosophy of neural networks. These nonparametric systems are composed of simple computational components, usually readily parallelizable, and often nonlinear. The components can be pieced together to tailor systems for specific applications. Graphical models for learning have these same features. Graphical models also have the expressibility of probabilistic knowledge representations that were developed in artificial intelligence to be used in knowledge acquisition contexts. They therefore form an important basis of knowledge refinement. Finally, graphical models for learning allow the powerful tools of statistics to be applied to the problem.

Once learning problems are specified in the common language of graphical models, their associated learning algorithms, their derivation, and their interrelationships can be explored. This allows commonalities between seemingly diverse pairs of algorithms—such as k-means clustering versus approximate methods for learning hidden Markov models, learning decision trees versus learning Bayesian networks (Buntine, 1991a), and Gibbs sampling versus the expectation maximization algorithm in Section 7.4—to be understood as variations of one another. The framework is important as an educational tool.

## Appendix A. Proofs of Lemmas and Theorems

### A.1 Proof of Theorem 2.1

A useful property of independence is that $A$ is independent of $B$ given $C$ if and only if $p(A, B, C) = f(A, C)g(B, C)$ for some functions $f$ and $g$. The only if result follows directly from this property. The proof of the if result makes use of the following simple lemma. If $A$ is independent of $B$ given $X - A - B$, and $p(X) = \prod_i f_i(X_i)$ for some functions $f_i > 0$ and variable sets $X_i \subseteq X$, then:

$$p(X) = \prod_i g_i(X_i - B)\, h_i(X_i - A) \tag{27}$$

for some functions $g_i, h_i > 0$. Notice that it is known $p(X) = g(X - B)\, h(X - A)$ for some functions $g$ and $h$ by independence. Instantiate the variables in $B$ to some value $b$. Then:

$$g(X - B)\, h(X - A, B = b) \;=\; \prod_i f_i(X_i, B = b)\,.$$

Similarly, instantiate $A$ to $a$, then:

$$g(X - B, A = a)\, h(X - A) \;=\; \prod_i f_i(X_i, A = a)\,.$$

Multiplying both sides of the two equalities together, and substitute in

$$g(X - B, A = a)\, h(X - A, B = b) \;=\; p(X, A = a, B = b) \;=\; \prod_i f_i(X_i, A = a, B = b)\,.$$

Get:

$$p(X) \prod_i f_i(X_i, A = a, B = b) \;=\; \prod_i f_i(X_i, B = b) \prod_i f_i(X_i, A = a)\,.$$

This is defined for all $X$ since the domain is a cross product. The lemma holds because all functions are strictly positive if

$$g_i(X_i - B) \;=\; \frac{f_i(X_i, B = b)}{f_i(X_i, B = b, A = a)}$$

and

$$h_i(X_i - A) \;=\; f_i(X_i, A = a)\,.$$

The final proof of the if result follows by applying Equation (27) repeatedly. Suppose the variables in $X$ are $x_1, \ldots, x_v$. Now $p(X) = f_0(X)$ for some strictly positive function $f_0$. Therefore:

$$p(X) \;=\; g_0(X - \{x_1\}) h_0(\{x_1\} \cup neighbors(x))\,.$$

Denote $A_{i,0} = X - \{x_i\}$ and $A_{i,1} = \{x_i\} \cup neighbors(x_i)$. Repeating the application of Equation (27) for each variable yields:

$$p(X) \;=\; \prod_{i_1 = 0,1} \cdots \prod_{i_v = 0,1} g_{i_1, \ldots, i_v}\Big(X - \bigcup_{j = 1, \ldots, v} A_{j, i_j}\Big)\,.$$

for strictly positive functions $g_{i_1, \ldots, i_v}$. Now, consider these functions. It is only necessary to keep the function $g_{i_1, \ldots, i_v}$ if the set $X - \bigcup_{j = 1, \ldots, v} A_{j, i_j}$ is maximal: it is not contained in any such set. Equivalently, keep the function $g_{i_1, \ldots, i_v}$ if the set $\bigcup_{j = 1, \ldots, v} A_{j, i_j}$ is minimal. The minimal such sets are the set of cliques on the undirected graph. The result follows.

## A.2 Proof of Theorem 6.1

It takes $O(|X|^2)$ operations to remove the deterministic nodes from a graph using Lemma 6.1. These nodes can be removed from the graph and then reinserted at the end. Hereafter, assume the graph contains no deterministic nodes. Also, denote the unknown variables in a set $Y$ by $unknown(Y) = Y - known(Y)$. Then, without loss of generality, assume that $X_i$ contains all known variables in the Markov blanket of $unknown(X_i)$.

Showing the independence model is equivalent amounts to showing for $i = 1, \ldots, P$ that $unknown(X_i)$ is independent of $\bigcup_{j \neq i} unknown(X_j)$ given $known(X)$. To test independence using the method of Frydenberg (1990), each plate must be expanded (that is, duplicate it the right number of times), moralize the graph, removing the given nodes, and then test for separability. The Markov blanket for each node in this expanded graph corresponds to those nodes directly connected in the moralized expanded graph. Suppose we have the finest unique partition $unknown(X_i)$ of the unknown nodes. $X_i$'s are then reconstructed by adding known variables in the Markov blankets for variables in $unknown(X_i)$. Suppose $V$ is an unknown variable in a plate, and $V_j$ are its instances once the plate is expanded. Now, by symmetry, every $V_j$ is either in the same element of the finest partition, or they are all in separate elements. If $V_j$ has a certain unknown variable in its Markov boundary outside the plate, then so must $V_k$ for $k \neq j$ by symmetry. Therefore $V_j$ and $V_k$ are in the same element of the partition. Hence by contradiction, if $V_j$ is in a separate element, that element occurs wholly within the plate boundaries. Therefore, this finest partition can be represented using plates, and the finest partition identified from the graph ignoring plates. The operation of finding the finest separated sets in a graph is quadratic in the size of the graph, hence the $O(|X|^2)$ complexity.

Assume the condition holds and consider Equation (23). Let $cliques(\tau)$ denote the subsets of variables in $\tau \cup parents(\tau)$ that form cliques in the graph formed by restricting $G$ to $\tau \cup parents(\tau)$ and placing an undirected arc between all parents. Let $\tau(X)$ be the set of chain components in $X$. From Frydenberg (1990), we have:

$$p(X|M) = \prod_{\tau \in \tau(X)} \prod_{i \in ind(\tau)} \prod_{C \in cliques(\tau)} g_C(C_i) .$$

Furthermore, if $u \in X_i$ is not known, then the variables in $u$'s Markov blanket will occur in $X_i$, and therefore, if $u \in C$ for some clique $C$, then $C \subseteq X_i$. Therefore cliques containing an unknown variable can be partitioned according to which subgraph they belong in. Let:

$$cliques'_j = \{C \ : \ C \in cliques(\tau(X)), \ unknown(X_j) \cap C \neq \emptyset\} ,$$

and add to this any remaining cliques wholly contained in the set so far:

$$cliques_j = cliques'_j \cup \left\{ C \ : \ C \in cliques(\tau(X)), \ C \subseteq \bigcup_{C' \in cliques'_j} C' \right\} .$$

Call any remaining cliques $cliques_0$. Therefore:

$$p(X|M) = \prod_{j=0}^{P} \prod_{C \in cliques_j} \prod_{i \in ind(C)} g_C(C_i) .$$

| Distribution | Functional form |
|---|---|
| $j \sim C$-dim multinomial$(\theta_1, \ldots, \theta_C)$ | $\theta_j$ for $j \in [1, \ldots, C]$ |
| $y\|x \sim$ Gaussian$(x^\dagger \theta, \sigma)$ | $\frac{1}{\sqrt{2\pi\sigma}} \exp\left(-\frac{1}{2\sigma}(y - x^\dagger \theta)^2\right)$ for $y \in \Re, x \in \Re^d$ |
| $x \sim$ Gamma$(\alpha > 0, \beta > 0)$ | $\frac{\beta^\alpha}{\Gamma(\alpha)} x^{\alpha-1} e^{-\beta x}$ for $x \in \Re^+$ |
| $\theta \sim C$-dim Dirichlet$(\alpha_1, \ldots, \alpha_C)$ | $\frac{1}{Beta(\alpha_1, \ldots, \alpha_C)} \prod_{i=1}^C \theta_i^{\alpha_i - 1}$ |
| $x \sim d$-dim Gaussian$(\mu \in \Re^d, \Sigma \in \Re^{d \times d})$ | $\frac{\det^{1/2}\Sigma}{(2\pi)^{d/2}} \exp\left(-\frac{1}{2}(x - \mu)^\dagger \Sigma (x - \mu)\right)$ for $x \in \Re^d$ |
| $S \sim d$-dim Wishart$(\alpha \geq d, \Sigma \in \Re^{d \times d})$ | $\frac{\det^{-\alpha/2}\Sigma \ \det^{\alpha - d - 1/2} S}{2^{d\alpha/2} \pi^{d(d-1)/4} \prod_{i=1}^d \Gamma((\alpha + 1 - i)/2)} \exp\left(-\frac{1}{2}\text{trace}\Sigma^{-1} S\right)$ |
| | for $S, \Sigma$ symmetric positive definite |

Table 2: Distributions and their functional form

Results in:

$$f_j(known(X_{j,*})) = \int_{unknown(X_j,*)} \prod_{C \in cliques_j} \prod_{i \in ind(C)} g_C(C_i) \, d \, unknown(X_j, *) \, .$$

Furthermore, the potential functions on the cliques in $G_i$ are well defined as described.

### A.3 Proof of Corollary 6.1.1

If $X_j = \tau_j \cup ndparents(\tau_j)$, then every clique in a chain component in $\tau_j$ will occur in $cliques_j$. Therefore:

$$\prod_{C \in cliques_j} \prod_{i \in ind(C)} g_C(C_i) = \prod_{\tau \in \tau_j} p(\tau | ndparents(\tau)) \, ,$$
$$= p(\tau_j | ndparents(\tau_j)) \, .$$

### A.4 Proof of Lemma 6.3

Consider the definition of the Markov blanket. If a directed arc is added between the nodes, then the Markov blanket will only change for an unknown node $X$ if $U$ now enters the set of non-deterministic parents of the chain-components containing non-deterministic children of $X$. This will not effect the subsequent graph separability, however, because it will only subsequently add arcs between $U$, a given node, and other nodes.

## Appendix B. The exponential family

The exponential family of distributions was described in Definition 4.1. The common use of the exponential family exists because of Theorem 4.1. Table 2 gives a few exponential family distributions and their functional form. Further details and more extensive tables can be found in most Bayesian textbooks on probability distributions (DeGroot, 1970; Bernardo & Smith, 1994). Table 3 gives some standard conjugate prior distributions for those in Table 2, and Table 4 gives their matching posteriors (DeGroot, 1970; Bernardo & Smith, 1994). For the distributions in Table 2 with priors in Table 3 , Table 5 gives their matching evidence derived using Lemma 6.4 and cancelling a few common terms.

| Distribution | Conjugate prior |
|---|---|
| $j \sim C$-dim multinomial | $\theta \sim$ Dirichlet$(\alpha_1, \ldots, \alpha_C)$ |
| $y\|x \sim$ Gaussian | $\theta\|\sigma \sim d$-dim Gaussian$(\theta_0, \frac{1}{\sigma^2}\Sigma_0)$, $\sigma^{-2} \sim$ Gamma$(\alpha_0/2, 2/\beta_0)$ |
| $x \sim$ Gamma | $\beta\|\alpha \sim$ Gamma$(\alpha_0, \beta_0)$ |
| $x \sim d$-dim Gaussian | $\mu\|\Sigma \sim$ Gaussian$(\mu_0, N_0\Sigma)$, $\Sigma \sim$ Wishart$(\delta_0, S_0)$ |

Table 3: Distributions and their conjugate priors

| Distribution | Conjugate posterior |
|---|---|
| $j \sim C$-dim multinomial | $\theta \sim$ Dirichlet$(n_1 + \alpha_1, \ldots, n_C + \alpha_C)$<br>for $n_c = \sum_{i=1}^N 1_{j_i=c} = \# < j's = c >$ |
| $y\|x \sim$ Gaussian | $\theta\|\sigma \sim d$-dim Gaussian$(\overline{\theta}, \frac{1}{\sigma^2}\Sigma)$, $\sigma^{-2} \sim$ Gamma$((\alpha_0 + N)2, 2/\beta)$,<br>for $\Sigma = \Sigma_0 + \sum_{i=1}^N y_i y_i^\dagger$, $\overline{\theta} = \Sigma^{-1}\left(\Sigma_0\theta_0 + \sum_{i=1}^N x_i y_i\right)$,<br>for $\beta = \sum_{i=1}^N (x_i - \overline{\theta}^\dagger y_i)^2 + (\overline{\theta} - \theta_0)^\dagger \Sigma_0(\overline{\theta} - \theta_0) + \beta_0$ |
| $x \sim$ Gamma | $\beta\|\alpha \sim$ Gamma$(N\alpha + \alpha_0, \sum_{i=1}^N x_i + \beta_0)$ |
| $x \sim d$-dim Gaussian | $\mu\|\Sigma \sim$ Gaussian$(\overline{\mu}, (N + N_0)\Sigma)$, $\Sigma \sim$ Wishart$(N + \delta_0, S + S_0)$<br>for $\overline{\mu} = \overline{x} + \frac{N_0}{N+N_0}(\mu_0 - \overline{x})$,<br>for $S = \sum_{i=1}^N (x_i - \overline{x})(x_i - \overline{x})^\dagger$ |

Table 4: Distributions and matching conjugate posteriors

| Distribution | Evidence |
|---|---|
| $j \sim C$-dim multinomial | $Beta(n_1 + \alpha_1, \ldots, n_C + \alpha_C)/Beta(\alpha_1, \ldots, \alpha_C)$ |
| $y\vert x \sim$ Gaussian | $\frac{\det^{1/2}\Sigma_0}{\pi^{N/2}\det^{1/2}\Sigma}\frac{\Gamma((\alpha_0+N)/2)\beta^{(\alpha_0+N)/2}}{\Gamma(\alpha_0/2)\beta_0^{\alpha_0/2}}$ |
| $x \sim$ Gamma | $\frac{\beta_0^{\alpha_0}\Gamma(N\alpha+\alpha_0)}{\Gamma(\alpha_0)\left(\sum_{i=1}^{N}x_i+\beta_0\right)^{N\alpha+\alpha_0}}$ for $\alpha$ fixed |
| $x \sim d$-dim Gaussian | $\frac{\det^{\delta_0/2}S_0}{(\pi)^{dN/2}\det^{(\delta_0+N)/2}(S+S_0)}\frac{N_0^d}{(N+N_0)^d}\prod_{i=1}^{d}\frac{\Gamma((\delta_0+N-1-i)/2)}{\Gamma((\delta_0-1-i)/2)}$ |

Table 5: Distributions and their evidence

In the case where the functions $w_i$ are *full rank* in $\theta$ (dimension of $\theta$ is $k$, same as $w$, and the Jacobian of $w$ with respect to $\theta$ is invertible, $\det\left(\frac{\mathrm{d}w(\theta)}{\mathrm{d}\theta}\right) \neq 0$), then various moments of the distribution can be easily found:

$$\mathcal{E}_{x\vert y,\theta}\left(t(x,y)\right) = \left(\frac{\mathrm{d}w(\theta)}{\mathrm{d}\theta}\right)^{-1}\frac{\mathrm{d}Z(\theta)}{\mathrm{d}\theta} . \qquad (28)$$

The vector function $w(\theta)$ now has an inverse and it is referred to as the *link function* (McCullagh & Nelder, 1989). This yields:

$$\mathcal{E}_{x\vert y,\theta}\left(\exp\left(\sum_{i=1}^{k}\phi_i\, t_i(x,y)\right)\right) = \frac{Z(w^{-1}(\phi+w(\theta)))}{Z(\theta)} . \qquad (29)$$

These are important because if the normalization constant $Z$ can be found in closed form, then it can be differentiated and divided, for instance, symbolically, to construct formula for various moments of the distribution such as $\mathcal{E}_{x\vert\theta}\left(t_i(x)\right)$ and $\mathcal{E}_{x\vert\theta}\left(t_i(x)t_j(x)\right)$. Furthermore, Equation (28) implies that derivatives of the normalization constant, $\mathrm{d}Z(\theta)/\mathrm{d}\theta$, can be found by estimating moments of the sufficient statistics (for instance, by Markov chain Monte Carlo methods).

## Acknowledgements

The general program presented here is shared by many, including Peter Cheeseman, who encouraged this development from its inception. These ideas were presented in formative stages at Snowbird 1993 (Neural Networks for Computing), April 1993, and to the Bayesian Analysis in Expert Systems (BAIES) group in Pavia, Italy, June 1993. Feedback from that group helped further develop these ideas. Graduate students at Stanford and Berkeley have also received various incarnations of this ideas. Thanks also to George John, Ronny Kohavi, Scott Schmidler, Scott Roy, Padhraic Smyth, and Peter Cheeseman for their feedback on drafts, and to the JAIR reviewers. Brian Williams pointed out the extension of the decomposition theorems to the deterministic case. Brian Ripley reminded me of the extensive features of S.

# References

Andersen, S., Olesen, K., Jensen, F., & Jensen, F. (1989). HUGIN—a shell for building Bayesian belief universes for expert systems. In *International Joint Conference on Artificial Intelligence*, Detroit. Morgan Kaufmann, pp. 1080–1085.

Azevedo-Filho, A., & Shachter, R. (1994). Laplace's method approximations for probabilistic inference in belief networks with continuous variables.. In de Mantaras, R. L., & Poole, D. (Eds.)(1994). *Uncertainty in Artificial Intelligence: Proceedings of the Tenth Conference*, Seattle, Washington, pp. 28–36.

Becker, R., Chambers, J., & Wilks, A. (1988). *The New S Language*. Pacific Grove, California: Wadsworth & Brooks/Cole.

Berger, J. (1985). *Statistical Decision Theory and Bayesian Analysis*. New York: Springer-Verlag.

Bernardo, J., & Smith, A. (1994). *Bayesian Theory*. Chichester: John Wiley.

Besag, J., York, J., & Mollie, A. (1991). Bayesian image restoration with two applications in spatial statistics. *Ann. Inst. Statist. Math., 43*(1), 1–59.

Box, G., & Tiao, G. (1973). *Bayesian Inference in Statistical Analysis*. Reading, Massachusetts: Addison-Wesley.

Breiman, L., Friedman, J., Olshen, R., & Stone, C. (1984). *Classification and Regression Trees*. Belmont, California: Wadsworth.

Bretthorst, G. (1994). An introduction to model selection using probability theory as logic. In Heidbreder, G. (Ed.). (1994), *Maximum Entropy and Bayesian Methods*, Kluwer Academic. (Proceedings, at Santa Barbara, 1993.)

Buntine, W. (1991a). Classifiers: A theoretical and empirical study. In *International Joint Conference on Artificial Intelligence*, Sydney, Morgan Kaufmann, pp. 638–644.

Buntine, W. (1991b). Learning classification trees. In Hand, D. (Ed.), *Artificial Intelligence Frontiers in Statistics*, London: Chapman & Hall, pp. 182–201.

Buntine, W. (1991c). Theory refinement of Bayesian networks. In D'Ambrosio, B., Smets, P., & Bonissone, P. (Eds.), *Uncertainty in Artificial Intelligence: Proceedings of the Seventh Conference* Los Angeles, California.

Buntine, W. (1994). Representing learning with graphical models. Technical Report FIA-94-14, Artificial Intelligence Research Branch, NASA Ames Research Center. Submitted.

Buntine, W., & Weigend, A. (1991). Bayesian back-propagation. *Complex Systems, 5*(1), 603–643.

Buntine, W., & Weigend, A. (1994). Computing second derivatives in feed-forward networks: a review. *IEEE Transactions on Neural Networks, 5*(3).

Casella, G., & Berger, R. (1990). *Statistical Inference*. Belmont, California: Wadsworth & Brooks/Cole.

Çinlar, E. (1975). *Introduction to Stochastic Processes*. Prentice Hall.

Chambers, J., & Hastie, T. (Eds.). (1992). *Statistical Models in S*. Pacific Grove, California: Wadsworth & Brooks/Cole.

Chan, B., & Shachter, R. (1992). Structural controllability and observability in influence diagrams.. In Dubois, D., Wellman, M., D'Ambrosio, B., & Smets, P. (Eds.), (1992). *Uncertainty in Artificial Intelligence: Proceedings of the Eight Conference*, Stanford, California, pp. 25–32.

Charniak, E. (1991). Bayesian networks without tears. *AI Magazine, 12*(4), 50–63.

Cheeseman, P., Self, M., Kelly, J., Taylor, W., Freeman, D., & Stutz, J. (1988). Bayesian classification. In *Seventh National Conference on Artificial Intelligence*, Saint Paul, Minnesota, pp. 607–611.

Cheeseman, P. (1990). On finding the most probable model. In Shrager, J., & Langley, P. (Eds.), *Computational Models of Discovery and Theory Formation*. Morgan Kaufmann.

Cohen, W. (1992). Compiling prior knowledge into an explicit bias. In *Ninth International Conference on Machine Learning*, Morgan Kaufmann, pp. 102–110.

Cooper, G., & Herskovits, E. (1992). A Bayesian method for the induction of probabilistic networks from data. *Machine Learning, 9*(4), 309–348.

Cowell, R. (1992). BAIES—a probabilistic expert system shell with qualitative and quantitative learning. Bernardo, In Bernardo, J., Berger, J., Dawid, A., & Smith, A. (Eds.). (1992). *Bayesian Statistics 4*. Oxford University Press, pp. 595–600.

Dagum, P., Galper, A., Horvitz, E., & Seiver, A. (1994). Uncertain reasoning and forecasting. *International Journal of Forecasting*. Submitted.

Dagum, P., & Horvitz, E. (1992). Reformulating inference problems through selective conditioning.. In Dubois, D., Wellman, M., D'Ambrosio, B., & Smets, P. (Eds.), (1992). *Uncertainty in Artificial Intelligence: Proceedings of the Eight Conference*, Stanford, California, pp. 49–54.

Dawid, A. (1979). Conditional independence in statistical theory. *SIAM Journal on Computing, 41*, 1–31.

Dawid, A. P. (1976). Properties of diagnostic data distributions. *Biometrics, 32*, 647–658.

Dawid, A., & Lauritzen, S. (1993). Hyper Markov laws in the statistical analysis of decomposable graphical models. *Annals of Statistics, 21*(3), 1272–1317.

Dean, T., & Wellman, M. (1991). *Planning and Control*. San Mateo, California: Morgan Kaufmann.

DeGroot, M. (1970). *Optimal Statistical Decisions*. McGraw-Hill.

Dempster, A., Laird, N., & Rubin, D. (1977). Maximum likelihood from incomplete data via the EM algorithm. *Journal of the Royal Statistical Society B, 39*, 1–38.

Duda, R., & Hart, P. (1973). *Pattern Classification and Scene Analysis*. New York: John Wiley.

Frydenberg, M. (1990). The chain graph Markov property. *Scandinavian Journal of Statistics, 17*, 333–353.

Geiger, D., & Heckerman, D. (1994). Learning Gaussian networks.. In de Mantaras, R. L., & Poole, D. (Eds.)(1994). *Uncertainty in Artificial Intelligence: Proceedings of the Tenth Conference*, Seattle, Washington, pp. 235–243.

Geman, D. (1990). Random fields and inverse problems in imaging. In Hennequin, P. (Ed.), *École d'Été de Probabilités de Saint-Flour XVIII - 1988*. Springer-Verlag, Berlin. In Lecture Notes in Mathematics, Volume 1427.

Geman, S., & Geman, D. (1984). Stochastic relaxation, Gibbs distributions, and the Bayesian relation of images. *IEEE Transactions on Pattern Analysis and Machine Intelligence, 6*, 721–741.

Gilks, W., Clayton, D., Spiegelhalter, D., Best, N., McNeil, A., Sharples, L., & Kirby, A. (1993a). Modelling complexity: applications of Gibbs sampling in medicine. *Journal of the Royal Statistical Society B, 55*, 39–102.

Gilks, W., Thomas, A., & Spiegelhalter, D. (1993b). A language and program for complex Bayesian modelling. *The Statistician, 43*, 169–178.

Gill, P. E., Murray, W., & Wright, M. H. (1981). *Practical Optimization*. San Diego: Academic Press.

Griewank, A., & Corliss, G. F. (Eds.). (1991). *Automatic Differentiation of Algorithms: Theory, Implementation, and Application*, Breckenridge, Colorado. SIAM.

Heckerman, D., Geiger, D., & Chickering, D. (1994). Learning Bayesian networks: The combination of knowledge and statistical data. Technical Report MSR-TR-94-09 (Revised), Microsoft Research, Advanced Technology Division. Submitted *Machine Learning Journal*.

Heckerman, D. (1991). *Probabilistic Similarity Networks*. MIT Press.

Henrion, M. (1990). Towards efficient inference in multiply connected belief networks. In Oliver, R., & Smith, J. (Eds.), *Influence Diagrams, Belief Nets and Decision Analysis*, pp. 385–407. Wiley.

Hertz, J., Krogh, A., & Palmer, R. (1991). *Introduction to the Theory of Neural Computation*. Addison-Wesley.

Howard, R. (1970). Decision analysis: perspectives on inference, decision, and experimentation. *Proceedings of the IEEE, 58*(5).

Hrycej, T. (1990). Gibbs sampling in Bayesian networks. *Artificial Intelligence, 46,* 351–363.

Jeffreys, H. (1961). *Theory of Probability* (third edition). Oxford: Clarendon Press.

Johnson, D., Papdimitriou, C., & Yannakakis, M. (1985). How easy is local search? In *FOCS'85*, pp. 39–42.

Kass, R., & Raftery, A. (1993). Bayes factors and model uncertainty. Technical Report #571, Department of Statistics, Carnegie Mellon University, PA. Submitted to *Jnl. of American Statistical Association*.

Kjæruff, U. (1992). A computational scheme for reasoning in dynamic probabilistic networks.. In Dubois, D., Wellman, M., D'Ambrosio, B., & Smets, P. (Eds.), (1992). *Uncertainty in Artificial Intelligence: Proceedings of the Eight Conference*, Stanford, California, pp. 121–129.

Kohavi, R. (1994). Bottom-up induction of oblivious, read-once decision graphs : Strengths and limitations.. In *Twelfth National Conference on Artificial Intelligence*.

Lange, K., & Sinsheimer, J. (1993). Normal/independent distributions and their applications in robust regression. *Journal of Computational and Graphical Statistics, 2*(2).

Langley, P., Iba, W., & Thompson, K. (1992). An analysis of Bayesian classifiers.. In *Tenth National Conference on Artificial Intelligence*, San Jose, California, pp. 223–228.

Lauritzen, S., Dawid, A., Larsen, B., & Leimer, H.-G. (1990). Independence properties of directed Markov fields. *Networks, 20,* 491–505.

Little, R., & Rubin, D. (1987). *Statistical Analysis with Missing Data*. New York: John Wiley and Sons.

Loredo, T. (1992). The promise of Bayesian inference for astrophysics. In Feigelson, E., & Babu, G. (Eds.), *Statistical Challenges in Modern Astronomy*. Springer-Verlag.

MacKay, D. (1992). A practical Bayesian framework for backprop networks. *Neural Computation, 4,* 448–472.

MacKay, D. (1993). Bayesian non-linear modeling for the energy prediction competition. Report Draft 1.2, Cavendish Laboratory, University of Cambridge.

Madigan, D., & Raftery, A. (1994). Model selection and accounting for model uncertainty in graphical models using Occam's window. *Journal of the American Statistical Association*. To appear.

McCullagh, P., & Nelder, J. (1989). *Generalized Linear Models* (second edition). Chapman and Hall, London.

McLachlan, G. J., & Basford, K. E. (1988). *Mixture Models: Inference and Applications to Clustering*. New York: Marcel Dekker.

Meilijson, I. (1989). A fast improvement to the EM algorithm on its own terms. *J. Roy. Statist. Soc. B, 51*(1), 127–138.

Minton, S., Johnson, M., Philips, A., & Laird, P. (1990). Solving large-scale constraint-satisfaction and scheduling problems using a heuristic repair method. In *Eighth National Conference on Artificial Intelligence*, Boston, Massachusetts, pp. 17–24.

Neal, R. (1993). Probabilistic inference using Markov chain Monte Carlo methods. Technical Report CRG-TR-93-1, Dept. of Computer Science, University of Toronto.

Nowlan, S., & Hinton, G. (1992). Simplifying neural networks by soft weight sharing. In Touretzky, D. (Ed.), *Advances in Neural Information Processing Systems 4 (NIPS*91)*. Morgan Kaufmann.

Oliver, J. (1993). Decision graphs – an extension of decision trees. In *Proceedings of the Fourth International Workshop on Artificial Intelligence and Statistics*, pp. 343–350. Extended version available as TR 173, Department of Computer Science, Monash University, Australia.

Pearl, J. (1988). *Probabilistic Reasoning in Intelligent Systems*. Morgan Kaufmann.

Poland, W. (1994). *Decision Analysis with Continuous and Discrete Variables: A Mixture Distribution Approach*. Ph.D. thesis, Department of Engineering Economic Systems, Stanford University, Stanford, California.

Press, S. (1989). *Bayesian Statistics*. New York: Wiley.

Quinlan, J. (1989). Unknown attribute values in induction. In Segre, A. (Ed.), *Proceedings of the Sixth International Machine Learning Workshop* Cornell, New York. Morgan Kaufmann.

Quinlan, J. (1992). *C4.5: Programs for Machine Learning*. Morgan Kaufmann.

Ripley, B. (1981). *Spatial Statistics*. New York: Wiley.

Ripley, B. (1987). *Stochastic Simulation*. John Wiley & Sons.

Rivest, R. (1987). Learning decision lists. *Machine Learning, 2*(3), 229–246.

Russell, S., Binder, J., & Koller, D. (1994). Adaptive probabilistic networks. Technical Report CSD-94-824, July 1994, University of California, Berkeley.

Selman, B., Levesque, H., & Mitchell, D. (1992). A new method for solving hard satisfiability problems.. In *Tenth National Conference on Artificial Intelligence*, San Jose, California, pp. 440–446.

Shachter, R. (1986). Evaluating influence diagrams. *Operations Research, 34*(6), 871–882.

Shachter, R. (1990). An ordered examination of influence diagrams. *Networks*, *20*, 535–563.

Shachter, R., Andersen, S., & Szolovits, P. (1994). Global conditioning for probabilistic inference in belief networks.. In de Mantaras, R. L., & Poole, D. (Eds.)(1994). *Uncertainty in Artificial Intelligence: Proceedings of the Tenth Conference*, Seattle, Washington, pp. 514–522.

Shachter, R., & Heckerman, D. (1987). Thinking backwards for knowledge acquisition. *AI Magazine*, *8*(Fall), 55–61.

Shachter, R., & Kenley, C. (1989). Gaussian influence diagrams. *Management Science*, *35*(5), 527–550.

Smith, A., & Spiegelhalter, D. (1980). Bayes factors and choice criteria for linear models. *Journal of the Royal Statistical Society B*, *42*(2), 213–220.

Spiegelhalter, D. (1993). Personal communication.

Spiegelhalter, D., Dawid, A., Lauritzen, S., & Cowell, R. (1993). Bayesian analysis in expert systems. *Statistical Science*, *8*(3), 219–283.

Spiegelhalter, D., & Lauritzen, S. (1990). Sequential updating of conditional probabilities on directed graphical structures. *Networks*, *20*, 579–605.

Srinivas, S., & Breese, J. (1990). IDEAL: A software package for analysis of influence diagrams. In Bonissone, P. (Ed.), *Proceedings of the Sixth Conference on Uncertainty in Artificial Intelligence* Cambridge, Massachusetts.

Stewart, L. (1987). Hierarchical Bayesian analysis using Monte Carlo integration: computing posterior distributions when there are many possible models. *The Statistician*, *36*, 211–219.

Tanner, M. (1993). *Tools for Statistical Inference* (Second edition). Springer-Verlag.

Thomas, A., Spiegelhalter, D., & Gilks, W. (1992). BUGS: A program to perform Bayesian inference using Gibbs sampling. Bernardo, In Bernardo, J., Berger, J., Dawid, A., & Smith, A. (Eds.). (1992). *Bayesian Statistics 4*. Oxford University Press, pp. 837–42.

Tierney, L., & Kadane, J. (1986). Accurate approximations for posterior moments and marginal densities. *Journal of the American Statistical Association*, *81*(393), 82–86.

Titterington, D., Smith, A., & Makov, U. (1985). *Statistical Analysis of Finite Mixture Distributions*. Chichester: John Wiley & Sons.

van Laarhoven, P., & Aarts, E. (1987). *Simulated Annealing: Theory and Applications*. Dordrecht: D. Reidel.

Vuong, T. (1989). Likelihood ratio tests for model selection and non-nested hypotheses. *Econometrica*, *36*, 307–333.

Werbos, P. J., McAvoy, T., & Su, T. (1992). Neural networks, system identification, and control in the chemical process industry. In White, D. A., & Sofge, D. A. (Eds.), *Handbook of Intelligent Control*, pp. 283–356. Van Nostrand Reinhold.

Wermuth, N., & Lauritzen, S. (1989). On substantive research hypotheses, conditional independence graphs and graphical chain models. *Journal of the Royal Statistical Society B, 51*(3).

Whittaker, J. (1990). *Graphical Models in Applied Multivariate Statistics*. Wiley.

Wolpert, D. (1994). Bayesian backpropagation over functions rather than weights. In Tesauro, G. (Ed.), *Advances in Neural Information Processing Systems 6 (NIPS*93)*. Morgan Kaufmann.